\documentclass{article}

\usepackage{microtype}
\usepackage{graphicx}
\usepackage{subfigure}
\usepackage{booktabs} % for professional tables

\usepackage{hyperref}

\usepackage[accepted]{icml2020_arxiv}

\usepackage{hyperref}
\usepackage{url}

\usepackage{graphicx}

\usepackage{algorithm}
\usepackage{algorithmic}

\usepackage{wrapfig}

\usepackage{cuted}

\usepackage{tikz}
\usepackage{pgfplots}
\usepgfplotslibrary{units,groupplots}
\pgfplotsset{compat=newest}
\usetikzlibrary{bayesnet}
\usetikzlibrary{arrows}
\usepackage{color}
\usepackage{graphicx}
\usepackage{caption}
\usetikzlibrary{backgrounds}
\usepackage{enumitem}
\usepackage[title]{appendix}
\usepackage{placeins}

\usepackage{graphicx}
\usepackage{booktabs}
\usepackage{pgfplots}
\usepackage{listings}
\pgfplotsset{compat=1.11}
\usepackage{stfloats}

\usepackage{pifont}% http://ctan.org/pkg/pifont
\newcommand{\cmark}{\ding{51}}%
\newcommand{\xmark}{\ding{55}}%

\usepackage{defs}
\usepackage{lipsum}
\usepackage{amsmath, amsthm, amssymb}
\usepackage[font=small,labelfont=bf]{caption}
\usepackage[colorinlistoftodos,textsize=tiny]{todonotes}
\usepackage{xargs}  
\usepackage{xcolor}

\newcommand{\barC}{\bar{C}}

\newcommand{\tC}{\tilde{C}}
\newcommand{\tX}{\tilde{X}}

\newcommand{\tV}{\tilde{V}}
\newcommand{\tH}{\tilde{H}}
\newcommand{\tcH}{\tilde{\cH}}

\renewcommand{\tx}{\tilde{x}}

\newcommand{\tv}{\tilde{v}}
\renewcommand{\th}{\tilde{h}}

\renewcommand{\algorithmicrequire}{\textbf{initialize:}}
\renewcommand{\algorithmicensure}{\textbf{ensure:}}
\usepackage{eqparbox}
\renewcommand{\algorithmiccomment}[1]{\hfill\eqparbox{COMMENT}{\# #1}}
\newcommand{\wk}{{(k)}}

\newcommand{\key}{{\text{key}}}

\icmltitlerunning{Robustifying Sequential Neural Processes}

\begin{document}

\twocolumn[
% \icmltitle{Robust Meta-Transfer Learning via Memory Reconstruction}
\icmltitle{Robustifying Sequential Neural Processes}

% \icmltitle{Recurrent Memory Reconstruction and Attention for Meta-Transfer Learning}



\icmlsetsymbol{equal}{*}

\begin{icmlauthorlist}
\icmlauthor{Jaesik Yoon}{sap}
\icmlauthor{Gautam Singh}{rucs}
\icmlauthor{Sungjin Ahn}{rucs,ruccs}
\end{icmlauthorlist}

\icmlaffiliation{sap}{SAP}
\icmlaffiliation{rucs}{Department of Computer Science, Rutgers University}
\icmlaffiliation{ruccs}{Rutgers Center for Cognitive Science}

\icmlcorrespondingauthor{Jaesik Yoon}{jaesik.yoon.kr@gmail.com}
\icmlcorrespondingauthor{Sungjin Ahn}{sjn.ahn@gmail.com}

\icmlkeywords{Machine Learning, ICML}

\vskip 0.3in
]



\printAffiliationsAndNotice{}  % leave blank if no need to mention equal contribution

\begin{abstract}
% checked grammar through grammarly
When tasks change over time, \textit{meta-transfer learning} seeks to improve the efficiency of learning a new task via both \textit{meta-learning} and \textit{transfer-learning}. While the standard attention has been effective in a variety of settings, we question its effectiveness in improving meta-transfer learning since the tasks being learned are dynamic and the amount of context can be substantially smaller. In this paper, using a recently proposed meta-transfer learning model, Sequential Neural Processes (SNP), we first empirically show that it suffers from a similar underfitting problem observed in the functions inferred by Neural Processes. However, we further demonstrate that unlike the meta-learning setting, the standard attention mechanisms are not effective in meta-transfer setting.~To resolve, we propose a new attention mechanism, Recurrent Memory Reconstruction (RMR), and demonstrate that providing an imaginary context that is recurrently updated and reconstructed with interaction is crucial in achieving effective attention for meta-transfer learning. Furthermore, incorporating RMR into SNP, we propose Attentive Sequential Neural Processes-RMR (ASNP-RMR) and demonstrate in various tasks that ASNP-RMR significantly outperforms the baselines. 
\end{abstract}

\section{Introduction}

A central challenge in machine learning is to improve learning efficiency.~Among such approaches are meta-learning~\citep{schmidhuber1987evolutionary,bengio1990learning} and transfer learning~\citep{pratt1993discriminability,pan2009survey}. Meta-learning aims to learn the learning process itself and thus enables efficient learning (e.g., from a small amount of data) while transfer learning allows efficient warm-start of a new task by transferring knowledge from previously learned tasks.

In many scenarios, these two problems are not separate but appear in a combined way. For example, to build a customer preference model monthly, we would like to build it by transferring knowledge from the models of the previous months instead of starting from scratch, because the general preference of a customer would not change much across months. However, due to the monthly preference shift, e.g., due to a seasonal change, we also need to efficiently learn from the new observations of the new month.

Sequential Neural Processes (SNP)~ \citep{snp} is a new probabilistic model class to resolve the above-mentioned \textit{meta-transfer learning} problem.~In SNP, meta-transfer learning is modeled as stochastic processes that change with time (thus, a stochastic process of stochastic processes). At each time-step, the stochastic process---or equivalently a \textit{task}---is meta-learned from the context. SNP represents each task by a latent state of the Neural Process (NP) and models the temporal dynamics of such latent states using a recurrent state-space model~\citep{hafner2018learning}. This enables SNP to transfer the knowledge of the previous tasks to learn a new task.

It is well-known that Neural Processes (NP) suffers from the \textit{underfitting} problem because all context observations are encoded with limited expressiveness (e.g., by sum or mean encoding) to satisfy the order-invariant property~\citep{wagstaff2019limitations}. In~\citet{anp}, the authors observe that query-dependent attention can substantially mitigate the problem and propose Attentive Neural Processes (ANP). Therefore, a crucial next question is whether SNP---which is partly based on NP for task-level meta-learning but also equipped with temporal-transfer---also suffers from underfitting, and if so, how we can resolve the problem.

\begin{figure}
  %\vspace{-8mm}
  \begin{center}
  \includegraphics[width=0.45\linewidth]{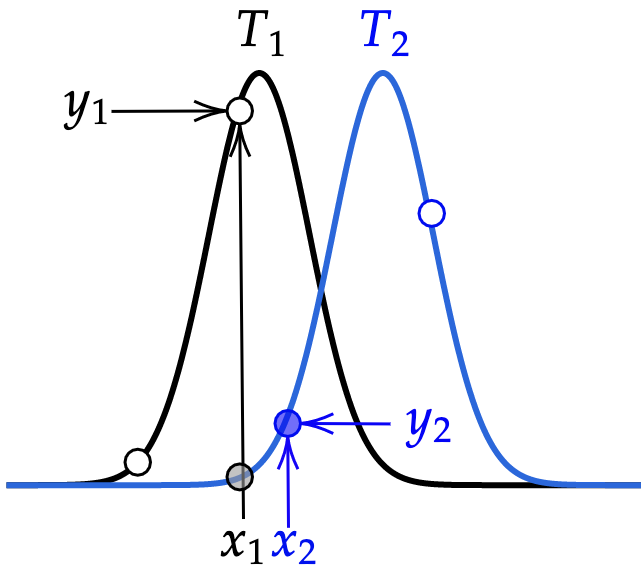}
  \end{center}
  \vspace{-4mm}
  \caption{Task shift. A context observation $c_1=(x_1,y_1)$ is provided for a task $T_1$ (black line). Then, the task is changed to $T_2$ (blue line). After the task is shifted to $T_2$, the value $f(x_2)$ is queried whose true value is $y_2$. A the standard attention will use the obsolete context $c_1 = (x_1,y_1)$ to infer $f(x_2)$ and a high attention-weight will be given to it because $x_1$ and $x_2$ are close. As a result, the attention will suggest a high value for $f(x_2)$ even though the true value $y_2$ is small. Our proposed model reconstructs $c_1$ so that its value can be properly adapted to the new task ($T_2$).}
\vspace{-3mm}
  \label{fig:dist_shift}
\end{figure}

In this paper, we argue not only that there is underfitting in SNP but also that it affects the robustness more severely in comparison to NP. We observe that this is because of two novel problems, \textit{sparse context} and \textit{obsolete context}, that occur in the novel setting of meta-transfer learning. In~\citet{snp}, it is shown that in comparison to meta-learning, meta-transfer learning is expected to learn a task more efficiently by using a much smaller amount of context or even empty context due to the availability of temporal transfer. However, this sparsity becomes an issue because when the context is small or empty, the attention---which is a remedy for underfitting---becomes highly ineffective or even not applicable. In the case of sparse context, it is possible to include the past contexts also for attention. However, this raises the second problem, the problem of obsolete context, because the past contexts come from tasks that are different from the current task. Thus, we argue that if the past contexts are not properly transformed for the current task, attention on them may hurt the performance. An illustration is shown in Fig.~\ref{fig:dist_shift}.

To this end, we propose a novel attention mechanism for meta-transfer learning, called Recurrent Memory Reconstruction and Attention (RMRA). In RMRA, to resolve the sparsity problem, for each task, we augment its original context with a generated \textit{imaginary} context. Thus, even if a task provides a small or empty context, we can still apply attention effectively on the generated imaginary context. To resolve the obsolete context problem, we generate this imaginary context using a novel \textit{Recurrent Memory Reconstruction} (RMR) mechanism. RMR temporally encodes \textit{all} the past context observations using recurrent updates and then reconstructs a reformed imaginary context for each task. In this way, the past contexts are properly transformed into a useful representation for the current task. In addition, we do not need to limit the attention to a finite-size window but can use the entire past information without storing it explicitly. By augmenting SNP with RMRA, we propose a novel and robust SNP model, called Attentive SNP-RMR.

Our main contributions are: (i) we identify why SNP should also suffer from underfitting and empirically show that it is indeed the case, (ii) we provide reasons why the existing attention mechanisms for NP are sub-optimal in the meta-transfer learning setting --- due to sparse and obsolete context --- and provide empirical analysis for it, (iii) we propose a novel Recurrent Memory Reconstruction and Attention (RMRA) mechanism to resolve the problem and we combine RMRA with SNP and thereby propose a novel model \textit{Attentive SNP-RMR}, and (iv) in our experiments, we empirically show that the RMRA mechanism resolves underfitting efficiently and effectively and results in superior performance in comparison to the baselines in various meta-transfer learning settings.

\section{Background}

\textbf{Neural Process (NP)}~\citep{garnelo2018neural} learns to learn a task $\tau$ to map an input $x \in \eR^{d_x}$ to an output $y \in \eR^{d_y}$ given a \textit{context} dataset $C = (X_C,Y_C) = \{(x^{(n)},y^{(n)})\}_{n\in[N_C]}$. Here, $N_C$ is the number of data points in $C$ and $[N_C]\equiv\{1, \ldots, N_C\}$. To learn a task distribution from this context, NP uses a distribution $P(z|C)$ to sample a task representation $z$. This makes NP a probabilistic meta-learning framework. Next, an observation model $p(y|x,z)$ takes an input $x$ and returns an output $y$. The generative process for NP conditioned on the context is given by:
\eq{
P(Y,z|X, C) = P(Y|X, z)P(z|C) 
} 
where $\smash{P(Y|X, z) = \prod_{n\in [N_D]} P(y^{(n)}|x^{(n)},z)}$ and $D = (X,Y)=\{(x^{(n)},y^{(n)})\}_{n\in[N_D]}$ is the \textit{target} dataset. To obtain the training data for this meta-learning setting, we draw multiple tasks from the true task distribution and sample $(C,D)$ for each task. Note that to implement NP, $C$ is encoded via a permutation-invariant function, such as $\sum_n \text{MLP}(x^{(n)}, y^{(n)})$. \citet{anp} argue that such a sum-aggregation produces an encoding that is not expressive enough and consequently hurts the observation model $P(Y|X, z)$. This is a key limitation of NP and is addressed by Attentive Neural Processes (ANP).

\textbf{Attentive Neural Process (ANP)}
~\citet{anp} identify the problem of \textit{underfitting} in NP. In underfitting, tasks learned from the context set fail to accurately predict the target points, including those in the context set. Also, the learned task distribution shows high uncertainty. Authors show that using a larger latent $z$ is also not sufficient.  Therefore to address this, ANP combines the observation model $p(y|x,z)$ with attention on the context points. This allows the model to take a query $x$ and perform attention on the most relevant data points to predict the target output $y$. To achieve this, an attention function $\text{Attend}(C;x^q)$ is implemented, which takes the context $C$ and a query $x^q$ and returns a read value $r_{x^q}$.
\eq{
r_{x^q} = \text{Attend}(C;x^q) = \sum_{n\in[N_C]} w_n y^{(n)}
}
Here, for each $y^{(n)} \in Y_C$, there is an associated weight $w_n$ which is computed using a similarity function, $\text{sim}(X_C, x^q)$. Using $r_{x}$, the observation model becomes $P(y|x,z,r_{x})$ and results in the following generative process.
\eq{
P(y,z|x, C)= P(y|x, z, r_{x})P(z|C)
}
where $(x,y)\in D$ is a target point. The ANP framework allows for the use of a variety of attentive mechanisms \citep{vaswani2017attention}.

\textbf{Sequential Neural Process (SNP)}~\citep{snp}
While the goal of meta-learning is to learn a single task distribution from a given context $C$, many situations consist of a sequence of tasks $\{\tau_t\}_{t\in[T]}$ that change gradually with time. Here, $t$ is a task-step. Thus, a framework that learns a task $\tau_t$ from a context $C_t$ must also utilize its similarity with the previous tasks to be able to learn with less context. Let $z_t$ denote a representation for a task $\tau_t$. Then SNP uses a distribution $p(z_t|z_{<t}, C_t)$ to learn from both $C_t$ and the previous task representations $z_{<t}$. From this standpoint, SNP is a meta-transfer learning framework. Given a task representation $z_t$, the model takes a query $x_t$ and generates an output $y_t$ through an observation model $p(y_t|x_t, z_t)$. 

For a task $\tau_t$, let $C_t = \{(x_t^{(n)},y_{t}^{(n)})\}$ be the context and let $(x_t, y_t)$ be a target point. Then, we describe the generative process for the target conditioned on the context as follows.
\eq{
P(Y,Z|X,C) = \pd{t}{T} P(y_t|x_t,z_t)P(z_t|z_{<t},C_t).
}
Here, $C$, $X$, $Y$ and $Z$ respectively denote the set aggregations of $C_t$, $x_t$, $y_t$ and $z_t$ over the entire roll-out.

\section{Attentive Meta-Transfer}
In this section, we describe our proposed model to resolve the problem of underfitting using attention in the setting of meta-transfer learning. We first propose a novel recurrent attention mechanism called Recurrent Memory Reconstruction (RMR), which resolves the problem of \textit{sparse context} that makes underfitting in SNP more severe and \textit{obsolete context} which turns the past contexts to noise due to task shift. We then propose to incorporate RMR into the SNP framework to obtain a robust meta-transfer learning model, Attentive Sequential Neural Processes-RMR (ASNP-RMR).

\subsection{Recurrent Memory Reconstruction}
\begin{algorithm}[t]
\footnotesize
\caption{Recurrent Memory Reconstruction}
\label{algo:rmr}
\begin{algorithmic}
    \ENSURE  $f_{xy}^c=$ MLP, \hspace{1mm} $f_{\text{order-invariant}}^c=\sum f_{xy}^c$ 
    \REQUIRE $h_0^x$, $\{ \th_0^{k} \}_{k\in [K]}$, $\tX_0$, $\{ \tv_0^{k} \}_{k\in [K]}$
    \FOR{$t\in[T]$}
    \STATE \# \textit{Context Processing}
    \STATE $r_t^c \leftarrow f_{\text{order-invariant}}^c(C_t)$
    \STATE $C_t \leftarrow \{(x_t^{(n)},f_{xy}^c(x_{t}^{(n)}, y_{t}^{(n)})) : (x_t^{(n)},y_{t}^{(n)}) \in C_t\}$
    \STATE
    \STATE \# \textit{Key Imagination}
    \STATE $h_t^x \leftarrow \text{RNN}(\tX_{t-1}, h_{t-1}^x, r_t^c)$
    \STATE $\tX_t \leftarrow f_X(h_t^x)$
    \STATE
    \STATE \# \textit{Value Imagination}
    \FOR{$k\in[K]$ in parallel}
        \STATE $\th_t^{(k)} \leftarrow \text{RNN}_k(\tx_t^{(k)}, \tv_{t-1}^{(k)}, \th_{t-1}^{(k)})$
    \ENDFOR
    \STATE $\{ \tv_t^{(k)} \}_{k\in[K]} \leftarrow \text{Attend}(C_t \cup \{(\tx_t^{(k)}, \th_t^{(k)})\}_{k\in[K]} ; \tX_t)$
    \STATE
    \STATE \# \textit{Store}
    \STATE $\tC_t \leftarrow \{(\tx_t^{(k)}, \tv_t^{(k)})\}_{k\in[K]}$
    \ENDFOR
\end{algorithmic}

\end{algorithm}
To resolve the limitations of the standard attention models in dealing with sparse and obsolete context, we propose the Recurrent Memory Reconstruction (RMR) mechanism. The key ideas in RMR are (i) to generate \textit{imaginary context} to complement the sparse context and (ii) to introduce \textit{recurrent reformation} to appropriately transform the obsolete context to a new useful representation upon a task-shift. The imaginary context is generated at each task-step using the updated representation of past contexts via recurrent reformation.

The main task of RMR at each task-step $t$ is to generate a new imaginary context $\tC_t$ from an encoded representation of the past contexts (both real and imaginary). The imaginary context contains $K$ memory cells, each of which is a key-value pair $(\tx_t^{(k)}, \tv_t^{(k)})$ for $k\in [K]\equiv \{1,\dots,K\}$. We denote the imagined key-set by $\smash{\tX_t = \{\tx_t^{(k)}\}_k}$ and the imagined value-set by $\smash{\tV_t = \{\tv_t^{(k)}\}_k}$. When generating imaginary context, we also use the real context $C_t$ gathered from the current task $\tau_t$  to inform the RMR about the characteristics of the current task. This is summarized by:
\eq{
\tC_t, \tcH_t = \text{RMR}(C_t, \tC_\tmo, \tcH_\tmo)
\label{eq:rmr}
}
where $\tcH_t = \tcH_t^\text{key} \cup \tcH_t^\text{val}$ is a set of RMR's hidden states, which consists of the hidden-states of key-generation and value-generation, respectively. RMR generates imaginary keys first and then, conditioning on the generated keys, the imaginary values.

\begin{figure*}
        \centering
        \includegraphics[width=1.0\linewidth]{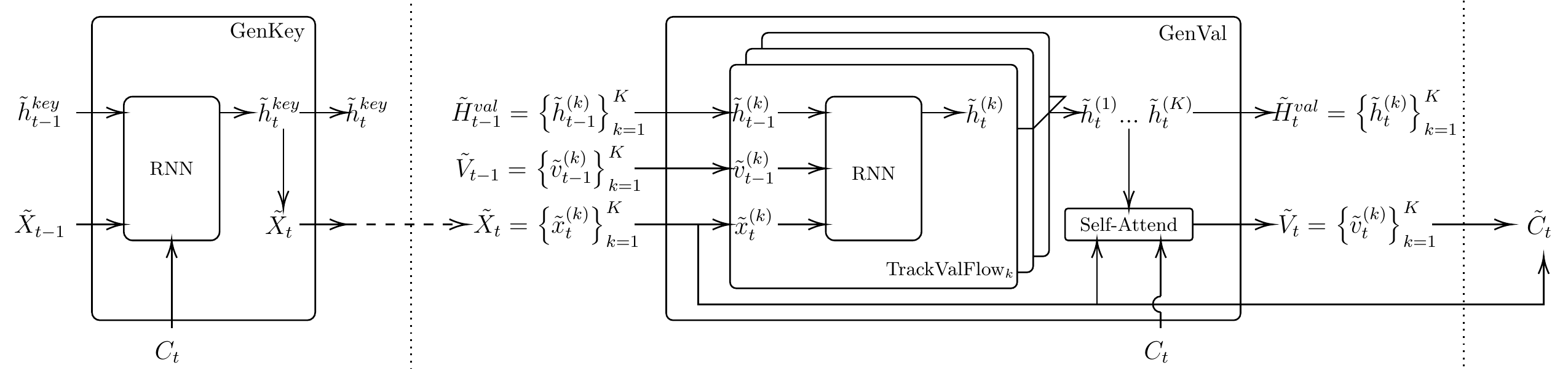}
\vspace{-5mm}
    \caption{Illustration of Recurrent Memory Reconstruction (RMR) for generating imaginary context.}
    \label{fig:rmr_pipe}
\vspace{-5mm}
\end{figure*}

\subsubsection{Generating Imaginary Keys} 
The main goal of imaginary key generation is to update the formation of the imaginary inputs (i.e., keys) upon a task-shift so that it can provide attention on a meaningful area of the new task. This requires the model to see (i) which input areas were tracked so far and to learn the task-shift dynamics via $\tX_\tmo$ and (ii) what the current task is via $C_t$. To this end, RMR deploys an RNN called \textit{key-tracker}. Therefore, the key-generation process becomes: 
\eq{
(\tX_t, \th_t^\key) = \text{GenerateKey}(C_t,\tX_\tmo,\th_\tmo^\key).
}
where $\th_t^\key$ is the hidden state of the key-tracker.

Specifically, we first concatenate the previous imaginary-keys  $\smash{\tx_\tmo^{(1)},\dots,\tx_\tmo^{(K)}}$ to make the input representation and then encode the real context $C_t$ using an order-invariant encoder $\smash{r_t = \sum_n \text{MLP}(x_t^{(n)}, y_t^{(n)})}$. We then provide the concatenated keys and the real context encoding to the key-tracker RNN as the inputs. The new imaginary key-set $\tX_t$ is generated from the updated hidden state $\th_t^\key$. To compute the order-invariant encoding, other implementation options are described in \citet{garnelo2018neural, eslami2018neural, gordon2019convolutional}.

\subsubsection{Generating Imaginary Values} 
Given the new imaginary keys, we generate the corresponding imaginary values. Similar to the key generation process, to generate a new value set $\tV_t$, RMR takes the new key $\tX_t$, the new real context $C_t$, and the previous value-set $\tV_\tmo$ as inputs. This results in the following summary of the value-generation process:
\eq{
\tV_t, \tH_t^\text{val} = \text{GenerateValue}(\tX_t, C_t, \tV_\tmo, \tH_\tmo^\text{val})
\label{eq:gen_val}
}
where $\tH_\tmo^\text{val}$ is a set of hidden states for value generation.

In designing the value generator, two principles play key roles. First, value-generation process should be aware of what has happened in the past to generate a useful value for the current task upon a task-shift. Second, the values in a value-set should be aware of each other in order to obtain an optimal formation of the values. To realize this, we implement the value generation by the following two components: value-flow tracking and value-flow interaction.

\textit{i) Value-Flow Tracking.} The purpose of this stage is to implement recurrence that captures value-transitions across the tasks. To this end, for each memory cell $k\in[K]$, we assign an RNN, TrackValueFlow updating:
\eq{
\th_t^{(k)} = \text{TrackValueFlow}(\tx_t^\wk,\tv_\tmo^\wk,\th_\tmo^\wk)
}
where $\th_t^{(k)}$ is the RNN hidden-state which we call the \textit{value-tracker state}. Here, each value-tracker state acts as a proposal for the final imaginary value, and each RNN is seen as \textit{tracking} the \textit{flow} of the value in a particular memory cell, hence the name.

\textit{ii) Value-Flow Interaction.} In the previous step, TrackValueFlow, each value was updated using the new key and its previous value. However, this step does not know what other values were generated. Hence, we require an interactive update. We revise each proposal value via self-attention on other values in the proposal value-set and the real context set. Specifically, we first use the proposal values to construct a proposal key-value set $\smash{\tC^\text{prop}_t = \{(\tx_{t}^{(k)}, \th_t^{(k)}\}_{k\in[K]}}$. We then combine it with the real context $C_t$ to get $C_t \cup \tC^\text{prop}_t$. Finally, we perform self-attention on this union using the imaginary keys as the attention queries and obtain the final imaginary value-set $\smash{\\tV_t=\{ \tv_t^{(k)} \}_{k\in[K]}}$:
\eq{
\tV_t = \text{Self-Attend}(C_t \cup \tC^\text{prop}_t ; \tX_t).
}
This completes the generation of imaginary context, $\tC_t = (\tX_t,\tV_t)$. Algorithm~\ref{algo:rmr} and Fig.~\ref{fig:rmr_pipe} show the described process of generating imaginary context.

\subsubsection{Reading in RMR}
To perform a read operation on the RMR at a given task-step $t$, we propose to perform attention on the extended context $\barC_t = C_t\cup \tC_t$. Given a query input $x_t^q$, we thus obtain the read value as follows $r_{x_t^q} = \text{Attend}(\barC_t; x_t^q)$.

\subsection{Attentive Sequential Neural Processes}
Using the imagined context obtained from RMR, we can now address the problems of sparse and obsolete context in SNP. In this section, we describe how we augment Sequential Neural Processes with RMR and, in doing so, propose \textit{Attentive Sequential Neural Processes-RMR}.

\subsubsection{Generative Process}
\begin{figure}
        \centering
        \includegraphics[width=\linewidth]{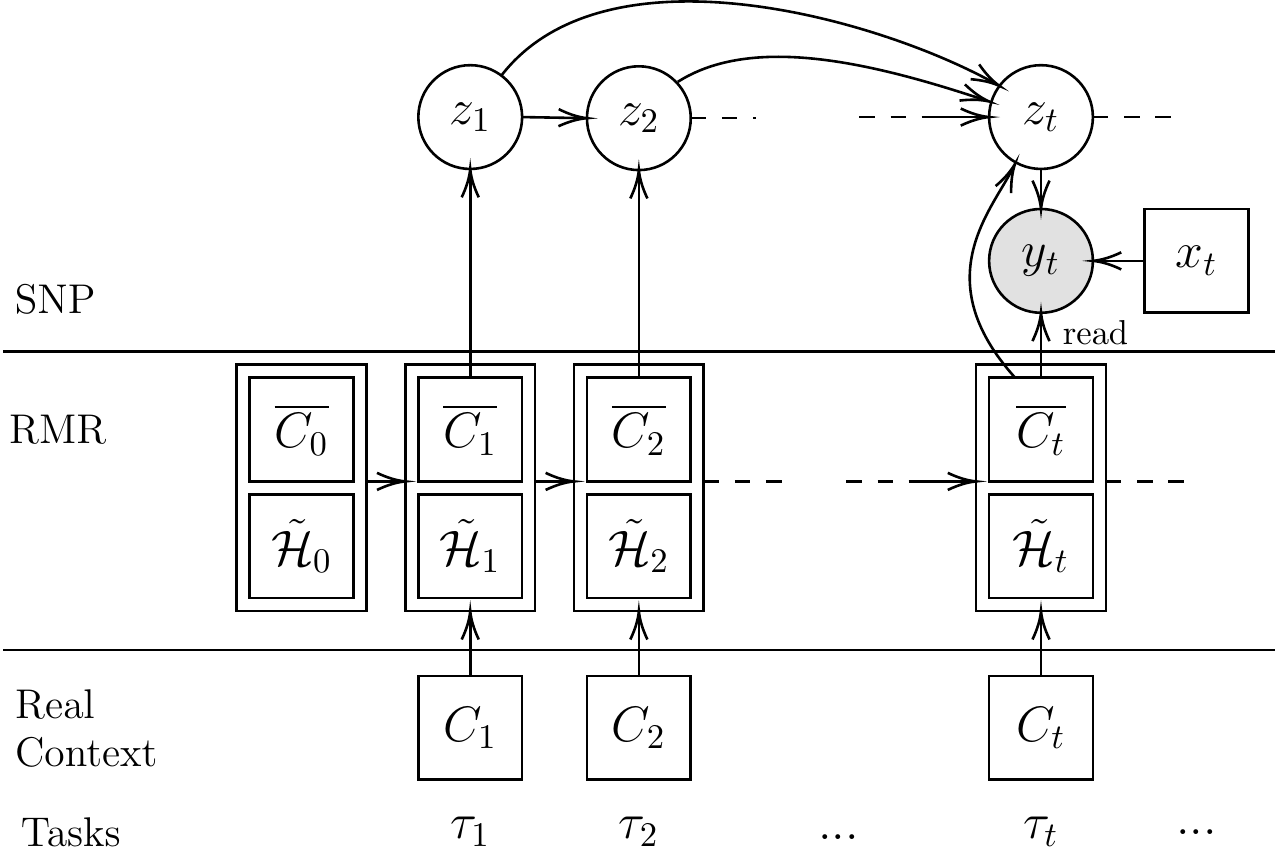}
\vspace{-4mm}
    \caption{Illustration of the generative process of ASNP-RMR. }
    \label{fig:asnp_pgm}
\vspace{-2mm}
\end{figure}

% To resolve underfitting along with sparse and obsolete context, 
We equip the observation model in SNP with the augmented memory $\barC_t$ provided by RMR. With this, the observation model becomes $P(y_t|x_t, z_t, \barC_t)$ (see~Fig. \ref{fig:asnp_pgm}). Then, given a target input $x_t$, the observation model reads $\barC_t$ to obtain an attention encoding $r_{x_t} = \text{Attend}(\barC_t; x_t)$.

In this way, combining SNP with RMR, the generative process is described as follows:
$P(Y,Z,\tC|X,C) = $
\eq{
\prod_{t=1}^T \underbrace{P(y_t|x_t,z_t,\barC_t)P(z_t|z_\lt,\barC_t)}_{\text{SNP with access to RMR}} \underbrace{P(\tC_t | \tC_\lt, C_{\leq t})}_{\text{RMR Update}}.\label{eq:gen}
}

We call this model Attentive Sequential Neural Process with RMR (ASNP-RMR). 

\subsubsection{Learning and Inference}\label{main_3.3}
Because the true posterior is not tractable, ASNP is trained using a variational approximation with the following auto-regressive formulation:
\eq{
P(Z|\tC,C,D) \approx \pd{t}{T}Q(z_t|z_\lt,\tC_{\leq t},C,D)
}
where $D = (X,Y)$ and $C$ is the set aggregations of $C_t$ over the entire roll-out.
For training, the following ELBO is maximized w.r.t. $\ta$ and $\phi$: $\cL_\text{ASNP}(\ta, \phi) = $
\begin{flalign}
\begin{aligned}
&\sm{t}{T} \eE_{Q_\phi(z_t|C,D)} \left[\log P_\ta(y_t|x_t,z_t,\tC_t,C_t)\right] \nn  \\
&- \eE_{Q_\phi(z_\lt)}\left[\KL ( Q_\phi(z_t|z_\lt, \tC_{\leq t},C,D) \parallel \right. \nn\\
&\left. \quad \quad P_\ta(z_t|z_\lt, \tC_{\leq t}, C_{\leq t}) )\right].
\label{eq:elbo}
\end{aligned}
\end{flalign}
For backpropagation, we use the reparameterization trick~\citep{vae}. See Appendix \ref{app:elbo_deri} for derivation.

%%%%%%%%%%%%%%%%%%%%%%%%%%%%%%%%%%%%%%%%%%%%%%%%%%%%%%%%%%%%%%%%%%%%%%%%%%%%%%%%%%%%%%%%
% related work / exp part
%%%%%%%%%%%%%%%%%%%%%%%%%%%%%%%%%%%%%%%%%%%%%%%%%%%%%%%%%%%%%%%%%%%%%%%%%%%%%%%%%%%%%%%%

\section{Related Works}
\label{sec:related_work}

Meta-learning approaches have become attractive for learning to learn new tasks at test-time.
In this line of work, GQN~\citep{eslami2018neural} renders 3D scenes from a few viewpoints, and NP~\citep{garnelo2018neural} generalizes GQN. Subsequently, ANP~\citep{anp} identifies and resolves the problem of underfitting in NP by using query-dependent representation. Similarly, in~\citet{rosenbaum2018learning}, the authors introduce attention to GQN to render complex 3D scenes in large procedurally-generated maps as in Minecraft. Functional Neural Processes (FNP)~\citep{louizos2019functional} learns a graph of dependencies between a pre-selected set of points and the training points for modeling distributions over functions. ANP-RNN~\citep{qin2019recurrent} encodes the target inputs via an RNN and use the hidden states as queries to attend on the context points in a meta-learning setting.
To extend NP to \textit{meta-transfer learning}, SNP~\citep{snp} models a sequence of tasks with sequential latent representations. It outperforms NP in  meta-transfer learning a sequence of tasks that come from different but related distributions. Recurrent Neural Process (RNP)~\citep{kumar2019spatiotemporal} also deals with meta-transfer learning, but it transfers via deterministic representations and shows high uncertainty with sparse context. \citet{willi2019recurrent} also propose a model named RNP for meta-transfer learning. 

The term \textit{meta-transfer learning} has also been used in connection to a different set of problems that deal with discovery of causal mechanisms~\citep{bengio2019meta}, fast-adaption from a large-scale trained model~\citep{sun2019meta}, and knowledge transfer to different architectures and tasks~\citep{jang2019learning}. In~\citet{kang2018transferable}, the authors propose transferable meta-learning to apply a meta-trained model to a task from a task distribution different from the trained.

\citet{santoro2018relational} and~\citet{goyal2019recurrent} propose approaches combining self-attention and recurrent update. Recurrent Memory Core (RMC)~\citep{santoro2018relational} self-attends the entire memory with a single input vector and updates the memory through an RNN module. Recurrent Independent Mechanisms (RIM)~\citep{goyal2019recurrent} recurrently updates the memory but self-attends only the memories that are estimated to be the most relevant using the visual input. Although these methods combine recurrent and attention modules, however unlike RMR, their memory elements and inputs contain only values. They do not deal with key-value pairs in a meta-transfer learning setting.

\section{Experiments}

In this section, we describe our experiments to answer two key questions:
\textit{i)} By resolving the problems of sparse context and obsolete context, can we improve meta-transfer learning? \textit{ii)} If yes, what are the needed memory sizes and computational overhead during training? We also perform an ablation on RMR to demonstrate the need for flow-tracking and flow-interaction. In the rest, we first describe the baselines and our experiment settings. We then describe our results on dynamic 1D regression, dynamic 2D image completion, and dynamic 2D image rendering.

\begin{table}[h]
\scriptsize
\centering
\begin{tabular}{@{}l|cccc@{}}
\toprule
         & \multicolumn{1}{l}{ANP} & \multicolumn{1}{l}{SNP} & \multicolumn{1}{l}{ASNP-W} & \multicolumn{1}{l}{ASNP-RMR} \\ \midrule
Sequential Latent      & \xmark                                   & \cmark                           & \cmark  & \cmark                     \\
Attention      & \cmark                                  & \xmark                            & \cmark & \cmark                                \\
Recurrent Memory   & \xmark                                  & \xmark                           & \xmark  & \cmark                \\\bottomrule
\end{tabular}
 \caption{Taxonomy of the models considered for evaluation.}
\label{tab:baseline_taxonomy}
\end{table}
\textbf{Baselines.}  We consider three baselines -- ANP, SNP and ASNP-W. These are characterized by whether or not they contain a sequential latent, attention mechanism or a recurrent memory (see Table~\ref{tab:baseline_taxonomy}). Among these, ASNP-W is an extension of SNP such that the observation model attends on a window of $K$-most recent contexts. Hence, ASNP-W contains sequential latent and attention but \textit{not} a recurrent memory. These baselines are to test whether sequential latent and the standard attention alone can solve underfitting without addressing the problems of sparse and obsolete context. We also test ANP and SNP using different latent sizes to investigate its effect on performance. Similarly, we test ASNP-RMR and ASNP-W using different memory sizes and analyze its effect. See Appendix~\ref{app:model_details} for more details on their implementation.

\begin{figure}
    \centering
    \begin{tikzpicture}[scale=0.38]
    \begin{groupplot}[group style={group size=2 by 2,horizontal sep = 40pt,vertical sep = 40pt}]
        \centering
        \nextgroupplot[
            width=9cm,height=7cm,
            scale only axis=true,
            title=\Large{Sparse-Context in Dynamic 1D},
            xlabel={Time-step},
            ylabel={Target-NLL},
            xmin=1, xmax=50,
            ymin=-1.2, ymax=2.1,
            xtick={1,6,11,16,21,26,31,36,41,46},
            ytick={-1.0,-0.5,0.0,0.5,1.0,1.5,2.0},
            xmajorgrids=true,
            ymajorgrids=true,
            legend pos=north east,
            grid style=dashed,
            label style={font=\Large},
            tick label style={font=\Large}  
        ]
        % ANP(h=128)
        \addplot[
            color=violet,
            mark=diamond,
            line width=1pt,
            mark size=2pt,
            ]
            coordinates { (1,1.9197051906585694)(4,0.9789312416315079)(7,0.4142063235887326)(10,0.1507407321431674)(13,0.01585308655165136)(16,-0.09870144122513011)(19,-0.1622086406708695)(22,-0.22636970155872405)(25,-0.28823553785448897)(28,-0.3336330035608262)(31,-0.37989111240953205)(34,-0.43533926852047444)(37,-0.47309858912602065)(40,-0.45589313872158527)(43,-0.34586695675738155)(46,-0.20773212666157634)(49,-0.1336690668638795)
            };
        % SNP(h=128)
        \addplot[
            color=green,
            mark=triangle,
            line width=1pt,
            mark size=2pt,
            ]
            coordinates { (1,1.9223112726211549)(4,1.183965950012207)(7,0.6421864368766547)(10,0.3779640135355294)(13,0.22469890556996688)(16,0.10355092300102114)(19,0.01643985846545547)(22,-0.0551476388622541)(25,-0.13271770982333692)(28,-0.18891274579335005)(31,-0.23275003502611072)(34,-0.29133679418824615)(37,-0.34431063583586363)(40,-0.3529680298827589)(43,-0.30979136885143815)(46,-0.26100859061116355)(49,-0.24894015351193957)
            };
        % SNPK(h=128,K=25)
        \addplot[
            color=red,
            mark=o,
            line width=1pt,
            mark size=2pt,
            ]
            coordinates {(1,1.9154291033744812)(4,0.9756521427631378)(7,0.3859184378013015)(10,0.14188254746142775)(13,-0.018340604561381042)(16,-0.16185474442027042)(19,-0.2534398017451167)(22,-0.3092644648416899)(25,-0.36490678440779445)(28,-0.40203095398843286)(31,-0.43003169558942317)(34,-0.44469965070486067)(37,-0.49966415330767633)(40,-0.48158973902463914)(43,-0.4346809407696128)(46,-0.39915483340620994)(49,-0.3427961060591042)
            };
        % ASNP(h=128,K=25)
        \addplot[
            color=blue,
            mark=square,
            line width=1pt,
            mark size=2pt,
            ]
            coordinates { (1,1.9129909467697144)(4,0.9779843148589135)(7,0.3168356418143958)(10,-0.09146490456303581)(13,-0.41183125554583966)(16,-0.6737789215054363)(19,-0.8363264921680093)(22,-0.9442536509037018)(25,-1.0093567198514939)(28,-1.0559395164251328)(31,-1.0960114961862564)(34,-1.1337012648582458)(37,-1.1566540223360062)(40,-1.1268552184104919)(43,-1.0795878905057907)(46,-1.0229877680540085)(49,-0.9684694135189056)
            };
        \nextgroupplot[
            width=9cm,height=7cm,
            scale only axis=true,
            title=\Large{Transfer-Prediction in Dynamic 1D},
            xlabel={Time-step},
            xmin=1, xmax=20,
            ymin=-1.4, ymax=-0.09,
            xtick={1,4,7,10,13,16,19},
            ytick={-1.3,-1.0,-0.7,-0.4,-0.1,0.2,0.5},
            xmajorgrids=true,
            ymajorgrids=true,
            grid style=dashed,
            label style={font=\Large},
            tick label style={font=\Large}  
        ]
        % ANP(h=128)
        \addplot[
            color=violet,
            mark=diamond,
            line width=1pt,
            mark size=2pt,
            ]
            coordinates { (1,-0.9416061650775372)(2,-1.2262835489213466)(3,-1.2742118388414383)(4,-1.2997202229499818)(5,-1.3113032710552215)(6,-1.31927170753479)(7,-1.3262271690368652)(8,-1.3293571436405183)(9,-1.329855695962906)(10,-1.330926548242569)(11,-1.3213062000274658)(12,-1.307245056629181)(13,-1.2871757793426513)(14,-1.2570760089159012)(15,-1.2142031091451644)(16,-1.1596074801683427)(17,-1.095600618124008)(18,-1.0232330107688903)(19,-0.9433265566825867)(20,-0.8576569068431854)
            };
        % SNP(h=128)
        \addplot[
            color=green,
            mark=triangle,
            line width=1pt,
            mark size=2pt,
            ]
            coordinates { (1,-0.12809800053015352)(2,-0.641817235276103)(3,-0.7781123042851686)(4,-0.8404379580914975)(5,-0.865075301527977)(6,-0.8797997981309891)(7,-0.8859508109092712)(8,-0.886740835905075)(9,-0.8668600162863731)(10,-0.8729201430082321)(11,-0.851157556772232)(12,-0.8273275870084763)(13,-0.8015621024370193)(14,-0.7663741889595985)(15,-0.7301100832223892)(16,-0.6923583444952964)(17,-0.6552526637911796)(18,-0.6167375686764717)(19,-0.5782664260268211)(20,-0.5377476958930493)
            };
        % SNPK(h=128,K=25)
        \addplot[
            color=red,
            mark=o,
            line width=1pt,
            mark size=2pt,
            ]
            coordinates {(1,-0.9827629655599595)(2,-1.2312241849303245)(3,-1.2461859893798828)(4,-1.2548606580495834)(5,-1.259203119277954)(6,-1.2544806599617004)(7,-1.2576628333330155)(8,-1.2589038717746734)(9,-1.2478355371952057)(10,-1.269187366962433)(11,-1.0802064448595048)(12,-1.0528197115659714)(13,-1.0090551209449767)(14,-0.952400284409523)(15,-0.8901846569776535)(16,-0.8219460982084275)(17,-0.750493391752243)(18,-0.6799005243182182)(19,-0.6089219073951244)(20,-0.5372977051138877)
            };
        % ASNP(h=128,K=25)
        \addplot[
            color=blue,
            mark=square,
            line width=1pt,
            mark size=2pt,
            ]
            coordinates { (1,-0.9090237456141039)(2,-1.2101020067930222)(3,-1.2765457516908645)(4,-1.3067767804861068)(5,-1.3242017579078675)(6,-1.3321631848812103)(7,-1.3378321528434753)(8,-1.3395658266544341)(9,-1.3391781103610993)(10,-1.3401928889751433)(11,-1.3356824505329132)(12,-1.3333749818801879)(13,-1.3274410355091095)(14,-1.3204669165611267)(15,-1.3118839299678802)(16,-1.3004737424850463)(17,-1.2869476616382598)(18,-1.2719783198833465)(19,-1.2554468202590943)(20,-1.2380664789676665)
            };
        \nextgroupplot[
            width=9cm,height=7cm,
            scale only axis=true,
            xlabel={Training Time (seconds in millions)},
            ylabel={Target-NLL},
            xmin=0, xmax=1,
            ymin=-0.8, ymax=0.7,
            xtick={0,0.1,0.2,0.3,0.4,0.5,0.6,0.7,0.8,0.9},
            ytick={-1.0,-0.5,0.0,0.5,1.0,1.5,2.0},
            xmajorgrids=true,
            ymajorgrids=true,
            legend pos=north east,
            grid style=dashed,
            label style={font=\Large},
            tick label style={font=\Large}  
        ]
        % ANP(h=128)
        \addplot[
            color=violet,
            mark=diamond,
            line width=1pt,
            mark size=2pt,
            ]
            coordinates {(0.0,3.879441022872925)(0.043137656517744066,-0.08343345671892166)(0.08710415742111206,0.09365399926900864)(0.1306301709766388,0.18979249894618988)(0.17464370837998391,0.32772597670555115)(0.21846891752958297,0.17359966039657593)(0.26195653024744986,-0.1061183512210846)(0.3055648998503685,-0.0953395813703537)(0.34874183187627794,-0.08874420821666718)(0.390960747569561,-0.13717295229434967)(0.43311534525203704,-0.08718615025281906)(0.47594420389127734,-0.17296069860458374)(0.5181123214154243,-0.11931850016117096)(0.5591702617154122,-0.1502402126789093)(0.6014285523591042,-0.25519058108329773)(0.6432521235210895,-0.24078744649887085)(0.685924524799347,0.10183477401733398)(0.7280001300938129,-0.10444195568561554)(0.770329166831255,-0.14752188324928284)(0.8135245993933677,0.08531724661588669)
            };
        % SNP(h=128)
        \addplot[
            color=green,
            mark=triangle,
            line width=1pt,
            mark size=2pt,
            ]
            coordinates {(0.0,3.7502639293670654)(0.01653523473238945,0.6202131509780884)(0.033716737009763714,0.6431546807289124)(0.051074106078624726,0.6024659872055054)(0.06844423023867607,0.6379345059394836)(0.08580669888901711,0.49957016110420227)(0.10315230814361573,0.2313992977142334)(0.12048361250901223,0.24092301726341248)(0.13781555161738396,0.17984244227409363)(0.15515578718996048,0.11136671155691147)(0.1720987511036396,0.15962418913841248)(0.1894449044225216,0.14264942705631256)(0.20678741455030442,0.037407755851745605)(0.22413739215493203,0.09264084696769714)(0.24149615135002137,-0.17757736146450043)(0.25885456029748916,-0.009743036702275276)(0.27620375875687597,0.4109596908092499)(0.29356171731948855,0.007579507771879435)(0.3109235620343685,-0.03988781198859215)(0.32829054717588424,0.22668132185935974)
            };
        % SNPK(h=128,K=25)
        \addplot[
            color=red,
            mark=o,
            line width=1pt,
            mark size=2pt,
            ]
            coordinates {(0.0,4.0455708503723145)(0.04604638727331162,-0.006093797739595175)(0.09282598804616927,0.06913607567548752)(0.14043421038866044,0.2329181432723999)(0.1868798249452114,0.23352986574172974)(0.23275507774448395,0.10236109793186188)(0.2788313188991547,-0.09015346318483353)(0.32504298880577087,-0.12333010137081146)(0.37127864724946025,-0.09493524581193924)(0.417497952947855,-0.22348536550998688)(0.4636252822380066,-0.16520541906356812)(0.510299887335062,-0.20305664837360382)
            };
        % ASNP(h=128,K=25)
        \addplot[
            color=blue,
            mark=square,
            line width=1pt,
            mark size=2pt,
            ]
            coordinates {(0.0,4.263464450836182)(0.07316226253461838,-0.4398621916770935)(0.14547069164037704,-0.5121957659721375)(0.21731221004128456,-0.5761966109275818)(0.29706557716083526,-0.35914328694343567)(0.37974421830916405,-0.39540550112724304)(0.4596941066687107,-0.6806992888450623)(0.5443719332129955,-0.6103744506835938)(0.6175325761225223,-0.5874729752540588)(0.689757723944664,-0.6668055057525635)(0.7639052141182423,-0.6171709895133972)(0.8362924066126347,-0.6359285712242126)(0.9117947164955139,-0.6945847272872925)(0.9875532632701397,-0.6814051866531372)(1.063239660141945,-0.7929672002792358)(1.13908621287632,-0.7186614871025085)(1.2152402589521407,-0.5418674945831299)(1.2926381221227645,-0.6622884273529053)(1.3681132526943685,-0.6839012503623962)(1.442671325302124,-0.4162028431892395)
            };
        \nextgroupplot[
            width=9cm,height=7cm,
            scale only axis=true,
            xlabel={Training Time (seconds in millions)},
            xmin=0, xmax=0.53,
            ymin=-1.4, ymax=-0.4,
            xtick={0,0.1,0.2,0.3,0.4,0.5},
            ytick={-1.3,-1.0,-0.7,-0.4,-0.1,0.2,0.5},
            xmajorgrids=true,
            ymajorgrids=true,
            grid style=dashed,
            legend cell align={left},
            legend style={font=\Large},
            label style={font=\Large},
            tick label style={font=\Large}  
        ]
        % ANP(h=128)
        \addplot[
            color=violet,
            mark=diamond,
            line width=1pt,
            mark size=2pt,
            ]
            coordinates { (0.0,1.7829561233520508)(0.01723185479545593,-0.9587297439575195)(0.034363292004585266,-1.109106421470642)(0.051521910090446474,-1.1086758375167847)(0.06887607977366447,-1.2110671997070312)(0.08608712583732606,-1.1564964056015015)(0.10418166974806786,-1.1632957458496094)(0.12307397210884094,-1.1846956014633179)(0.1423295143458843,-1.1963547468185425)(0.16156072795510293,-1.1980098485946655)(0.1807663339612484,-1.2334394454956055)(0.19974408896303178,-1.2304613590240479)(0.21921429082274438,-1.2348101139068604)(0.23873411834526062,-1.1642229557037354)(0.2583844015474319,-1.2680370807647705)(0.2774067032933235,-1.2588731050491333)(0.29686940248179433,-1.1639765501022339)(0.3164773274860382,-1.210078477859497)(0.33587506875109674,-1.2139290571212769)(0.3541020382192135,-1.2234989404678345)
            };
        % SNP(h=128)
        \addplot[
            color=green,
            mark=triangle,
            line width=1pt,
            mark size=2pt,
            ]
            coordinates {(0.0,1.6131079196929932)(0.00706634379863739,0.24884556233882904)(0.013924338387727737,0.1349499523639679)(0.020837400908231735,-0.04575634002685547)(0.027756824890613555,-0.19721931219100952)(0.034642638087749485,-0.31598326563835144)(0.041547724663496016,-0.2800004184246063)(0.04846317168140411,-0.4966962933540344)(0.05538094968438149,-0.38752880692481995)(0.06229636522531509,-0.6383458375930786)(0.06914463047838211,-0.5031114816665649)(0.07598361477398873,-0.48703500628471375)(0.08295629427027702,-0.6313153505325317)(0.08979986462450028,-0.7004234194755554)(0.0966407087750435,-0.8296911120414734)(0.10348178651690483,-0.6816088557243347)(0.11032640777516366,-0.45135942101478577)(0.11714794498252869,-0.7953423261642456)(0.12396201967072487,-0.7971197962760925)(0.13078889393401147,-0.8109719157218933)
            };
        % SNPK(h=128,K=25)
        \addplot[
            color=red,
            mark=o,
            line width=1pt,
            mark size=2pt,
            ]
            coordinates {(0.0,3.569810628890991)(0.019407648727178574,-0.5693945288658142)(0.03886005108165741,-0.6452805399894714)(0.05832308141303062,-0.7745949625968933)(0.077783045129776,-0.6274843215942383)(0.09743711151218415,-0.6388300061225891)(0.11697927942943573,-0.8197736740112305)(0.13651835760259629,-0.8835344314575195)(0.15605368117928506,-0.8977980613708496)(0.1755473922932148,-1.001202940940857)(0.1941073057358265,-1.0146583318710327)(0.21270117019724846,-0.8194541931152344)(0.23143715449166297,-0.9383155703544617)(0.25027186954832076,-1.0221847295761108)(0.26928488848471643,-1.0665501356124878)(0.28841267803812026,-1.101410150527954)(0.30744110044026374,-0.924191415309906)(0.32642194107055666,-1.08742094039917)(0.34539031123900416,-1.009696125984192)(0.36436438809108734,-1.0244301557540894)
            };
        % ASNP(h=128,K=25)
        \addplot[
            color=blue,
            mark=square,
            line width=1pt,
            mark size=2pt,
            ]
            coordinates {(0.0,1.7733650207519531)(0.02717477209496498,-1.0766754150390625)(0.054335284283638,-1.2273982763290405)(0.08215076549053192,-1.1798293590545654)(0.10969176513290406,-1.285051941871643)(0.13717227559733391,-1.2592777013778687)(0.16474404176831245,-1.2254877090454102)(0.19239891684150695,-1.282194972038269)(0.21989741796875,-1.2619869709014893)(0.24715660944890977,-1.2660483121871948)(0.2744979178521633,-1.2905672788619995)(0.30196030167078974,-1.3039449453353882)(0.3291427448050976,-1.294538140296936)(0.3563451492624283,-1.234697937965393)(0.3833075628540516,-1.3144865036010742)(0.4102862956328392,-1.3101783990859985)(0.43707419204473497,-1.2650789022445679)(0.4637435576069355,-1.2683517932891846)(0.4903503782515526,-1.276721477508545)(0.5173049025464058,-1.2942010164260864)
            };
            \legend{ANP,SNP,ASNP-W (K=25),ASNP-RMR (K=25)}
    \end{groupplot}    
    \end{tikzpicture}
    
    \vspace{-2mm}
    \captionof{figure}{\textit{Top:} Target-NLL computed at each time-step in the task sequence. \textit{Bottom:} Target-NLL convergence against the wall-clock time computed during training on a held-out set. Hidden unit size in all models is $128$.}
    \label{fig:1d_reg_graph}
    \vspace{-3mm}
\end{figure}
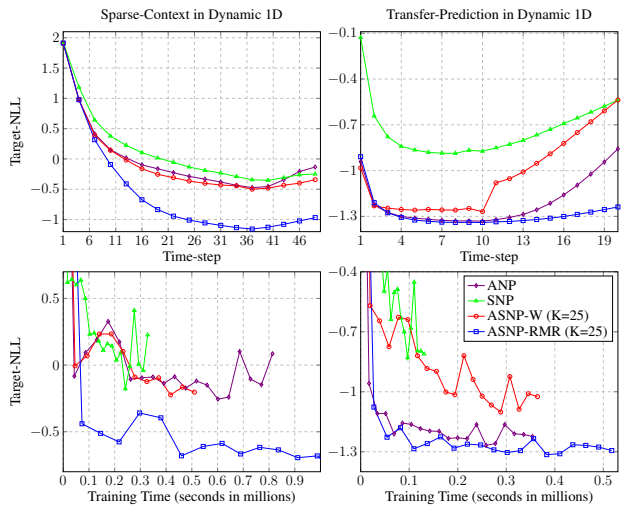

\textbf{Performance Metric and Context Regimes.} We evaluate our models on hundreds of held-out sequences of tasks and analyze their performance in modeling the target outputs. Our performance metric is \textit{Target-NLL} defined as $-\eE_{z\sim P(Z|C)} \log P(Y|X,Z,C)$. For dynamic 1D regression and dynamic 2D image completion, we consider two regimes for evaluation -- \textit{i)} \textit{Sparse-Context Regime.} In this regime, we consider task-sequences of length $50$. We provide a sparse context for $45$ randomly chosen tasks and empty context for the remaining. This regime tests the model's ability to transfer-learn from sparse contexts gathered from \textit{different} tasks. At every increment of the task-step, we expect the model's performance to improve as the model collects more context. \textit{ii)} \textit{Transfer-Prediction Regime.} In this regime, we consider task-sequences of length $20$. We provide a large-sized context for the first $10$ task-steps and empty context for the remaining. This regime allows the model to infer the first $10$ tasks and their dynamics with high certainty. Subsequently, the model must use this information to transfer-learn the remaining $10$ tasks. At every increment of the task-step, we expect the performance of any model to deteriorate as the contexts become more obsolete. However, if a model can deal with this problem, we expect it to show less degradation. 

For 2D dynamic image rendering also, we experiment on the sparse-context and transfer-prediction regimes but with task-sequences of length $6$. See Appendix \ref{app:task_details} for details about sequence lengths and context sizes.

\subsection{Dynamic 1D Regression}
In this setting, the tasks are 1D functions that change with time. To generate this dataset, we draw a function from a GP at each task-step such that the kernel-parameters of the GP change according to some linear dynamics. Hence, the model must estimate the function at unseen points and also track the shifts in its shape. We used a squared-exponential kernel for GPs. See Appendix \ref{app:gp_details} for more details.

\begin{figure}
    \centering
    \includegraphics[width=1.0\linewidth]{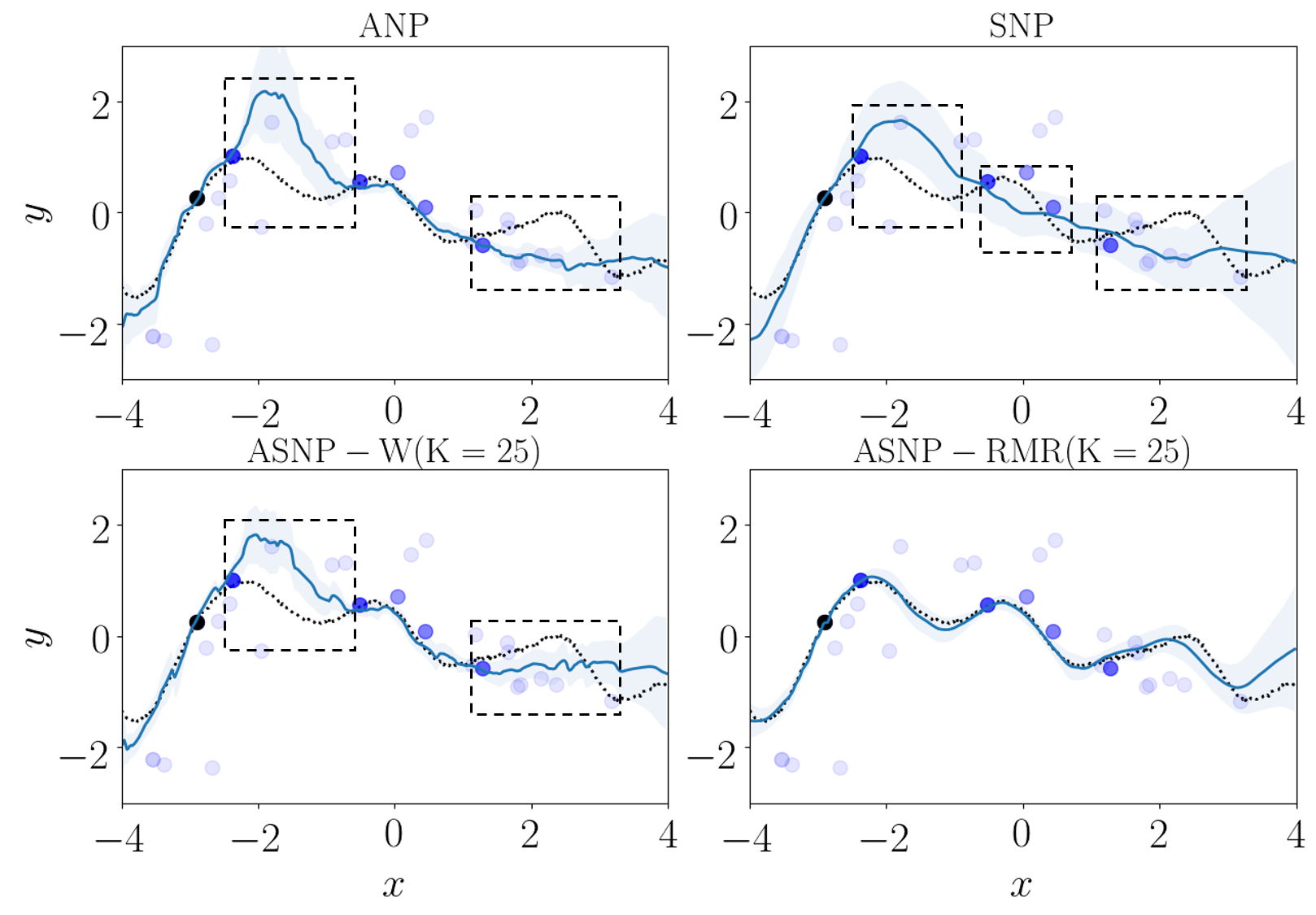}

    \vspace{-3mm}
    \captionof{figure}{Samples for dynamic 1D regression in sparse-context regime at $t=33$. \textit{Dotted line:} True function. \textit{Blue line:} Predicted function. \textit{Shaded light-blue region:} Uncertainty. \textit{Black dot:} Context points at $t=33$. \textit{Blue dots:} Past context points. Darker dots are more recent. \textit{Dashed-rectangles:} Regions where the predictions are inaccurate.}
    \label{fig:1d_reg_sam}
    \vspace{-3mm}
\end{figure}
\begin{figure*}
    \centering
    \begin{tikzpicture}[scale=0.34]
    \begin{groupplot}[group style={group size=3 by 1,horizontal sep = 64pt}]
        \centering
        \nextgroupplot[
            width=9cm,height=7cm,
            scale only axis=true,
            title=\huge{Dynamic 1D},
            xlabel={Memory size},
            ylabel={TargetNLL},
            xmin=1, xmax=4,
            xticklabels={9,25,100,$\infty$},
            ymin=-0.83, ymax=0.11,
            xtick={1,2,3,4},
            ytick={-0.8,-0.6,-0.4,-0.2,0.0},
            xmajorgrids=true,
            ymajorgrids=true,
            grid style=dashed,
            legend style={font=\large},
            legend cell align={left},
            tick label style={font=\huge},  
            label style={font=\huge},
        ]
        % ANP(h=128)
        \addplot[
            color=violet,
            mark=diamond,
            line width=1.5pt,
            mark size=2pt,
            ]
            coordinates {(1,-0.08913745731115341)(2,-0.08913745731115341)(3,-0.08913745731115341)(4,-0.08913745731115341)
            };
        % ANP(h=512)
        \addplot[
            color=violet,
            dashed,
            mark options={solid},
            mark=diamond,
            line width=1.5pt,
            mark size=2pt,
            ]
            coordinates {(1,-0.2553137540817261)(2,-0.2553137540817261)(3,-0.2553137540817261)(4,-0.2553137540817261)
            };
        % SNP(h=128)
        \addplot[
            color=brown,
            mark=triangle,
            line width=1.5pt,
            mark size=2pt,
            ]
            coordinates {(1,0.041198406368494034)(2,0.041198406368494034)(3,0.041198406368494034)(4,0.041198406368494034)
            };
        % SNP(h=512)
        \addplot[
            color=brown,
            dashed,
            mark options={solid},
            mark=triangle,
            line width=1.5pt,
            mark size=2pt,
            ]
            coordinates {(1,-0.1108688152115792)(2,-0.1108688152115792)(3,0.-0.1108688152115792)(4,-0.1108688152115792)
            };
        % SNP(h=1024)
        \addplot[
            color=brown,
            dotted,
            mark options={solid},
            mark=triangle,
            line width=1.5pt,
            mark size=2pt,
            ]
            coordinates {(1,-0.14447751641273499)(2,-0.14447751641273499)(3,-0.14447751641273499)(4,-0.14447751641273499)
            };
        % ASNPW
        \addplot[
            color=red,
            mark=o,
            line width=1.5pt,
            mark size=2pt,
            ]
            coordinates {(1,-0.05296846851706505)(2,0.04210847616195679)(3,-0.3826119601726532)(4,-0.26493313908576965)
            };
        % ASNP-RMR
        \addplot[
            color=blue,
            mark=square,
            line width=1.5pt,
            mark size=2pt,
            ]
            coordinates {(1,-0.4732550084590912)(2,-0.7188078165054321)(3,-0.814813369512558)
            };
        \nextgroupplot[
            width=9cm,height=7cm,
            scale only axis=true,
            title=\huge{Moving MNIST},
            xlabel={Memory size},
            xmin=1, xmax=4,
            xticklabels={9,25,100,$\infty$},
            ymin=-1.115, ymax=-0.89,
            xtick={1,2,3,4},
            ytick={-1.1,-1.0,-0.9,-0.85},
            xmajorgrids=true,
            ymajorgrids=true,
            legend pos=north east,
            grid style=dashed,
            legend cell align={left},
            label style={font=\huge},
            tick label style={font=\huge}  
        ]
        % ANP(h=128)
        \addplot[
            color=violet,
            mark=diamond,
            line width=1.5pt,
            mark size=2pt,
            ]
            coordinates {(1,-0.8975371265411377)(2,-0.8975371265411377)(3,-0.8975371265411377)(4,-0.8975371265411377)
            };
        % ANP(h=512)
        \addplot[
            color=violet,
            dashed,
            mark options={solid},
            mark=diamond,
            line width=1.5pt,
            mark size=2pt,
            ]
            coordinates {
            };
        % SNP(h=128)
        \addplot[
            color=brown,
            mark=triangle,
            line width=1.5pt,
            mark size=2pt,
            ]
            coordinates {(1,-0.9761032462120056)(2,-0.9761032462120056)(3,-0.9761032462120056)(4,-0.9761032462120056)
            };
        % SNP(h=512)
        \addplot[
            color=brown,
            dashed,
            mark options={solid},
            mark=triangle,
            line width=1.5pt,
            mark size=2pt,
            ]
            coordinates {(1,-0.9877805499732494)(2,-0.9877805499732494)(3,-0.9877805499732494)(4,-0.9877805499732494)
            };
        % SNP(h=1024)
        \addplot[
            color=brown,
            dotted,
            mark options={solid},
            mark=triangle,
            line width=1.5pt,
            mark size=2pt,
            ]
            coordinates {(1,-1.0040416532754899)(2,-1.0040416532754899)(3,-1.0040416532754899)(4,-1.0040416532754899)
            };
        % ASNPW
        \addplot[
            color=red,
            mark=o,
            line width=1.5pt,
            mark size=2pt,
            ]
            coordinates {(1,-0.9827292895317078)(2,-0.986812584400177)(3,-0.9923166477680206)(4,-0.957599088549614)
            };
        % ASNP-RMR
        \addplot[
            color=blue,
            mark=square,
            line width=1.5pt,
            mark size=2pt,
            ]
            coordinates {(1,-1.0560057806968688)(2,-1.0734858417510986)(3,-1.1031576871871949)
            };
        \nextgroupplot[
            width=9cm,height=7cm,
            scale only axis=true,
            title=\huge{Moving CelebA},
            xlabel={Memory size},
            xmin=1, xmax=4,
            xticklabels={9,25,100,$\infty$},
            ymin=-3.05, ymax=-2.05,
            xtick={1,2,3,4},
            ytick={-3.2,-3.0,-2.8,-2.6,-2.4,-2.2},
            xmajorgrids=true,
            ymajorgrids=true,
            legend style={at={(axis cs:4.6,-2.2)},anchor=north west},
            grid style=dashed,
            legend style={font=\LARGE, inner sep=4mm},
            legend cell align={left},
            label style={font=\huge},
            tick label style={font=\huge}  
        ]
        % ANP(h=128)
        \addplot[
            color=violet,
            mark=diamond,
            line width=1.5pt,
            mark size=2pt,
            ]
            coordinates {(1,-2.1074730050563812)(2,-2.1074730050563812)(3,-2.1074730050563812)(4,-2.1074730050563812)
            };\addlegendentry{ANP $(h=128)$}
        % ANP(h=512)  to make legend
        \addplot[
            color=violet,
            dashed,
            mark options={solid},
            mark=diamond,
            line width=1.5pt,
            mark size=2pt,
            ]
            coordinates {(1,-1.1074730050563812)(2,-1.1074730050563812)(3,-1.1074730050563812)(4,-1.1074730050563812)
            };\addlegendentry{ANP $(h=512)$}
        % SNP(h=128)
        \addplot[
            color=brown,
            mark=triangle,
            line width=1.5pt,
            mark size=2pt,
            ]
            coordinates {(1,-2.517590799331665)(2,-2.517590799331665)(3,-2.517590799331665)(4,-2.517590799331665)
            };\addlegendentry{SNP $(h=128)$}
        % SNP(h=512)
        \addplot[
            color=brown,
            dashed,
            mark options={solid},
            mark=triangle,
            line width=1.5pt,
            mark size=2pt,
            ]
            coordinates {(1,-2.71146052995324134)(2,-2.71146052995324134)(3,-2.71146052995324134)(4,-2.71146052995324134)
            };\addlegendentry{SNP $(h=512)$}
        % SNP(h=1024)
        \addplot[
            color=brown,
            dotted,
            mark options={solid},
            mark=triangle,
            line width=1.5pt,
            mark size=2pt,
            ]
            coordinates {(1,-2.7332194685935973)(2,-2.7332194685935973)(3,-2.7332194685935973)(4,-2.7332194685935973)
            };\addlegendentry{SNP $(h=1024)$}
        % ASNPW
        \addplot[
            color=red,
            mark=o,
            line width=1.5pt,
            mark size=2pt,
            ]
            coordinates {(1,-2.561289703845978)(2,-2.447293509244919)(3,-2.334347664117813)(4,-2.6233061933517456)
            };\addlegendentry{ASNP-W}
        % ASNP-RMR
        \addplot[
            color=blue,
            mark=square,
            line width=1.5pt,
            mark size=2pt,
            ]
            coordinates {(1,-2.805612049102783)(2,-2.9599591422080995)(3,-3.000567808151245)
            };\addlegendentry{ASNP-RMR}
    \end{groupplot}
    \end{tikzpicture}
    \vspace{-3mm}
\caption{Target-NLL computed as a function of memory sizes in sparse-context regime. Here, $h$ denotes the latent size.}
\label{fig:k_fig}
\vspace{-4mm}
\end{figure*}
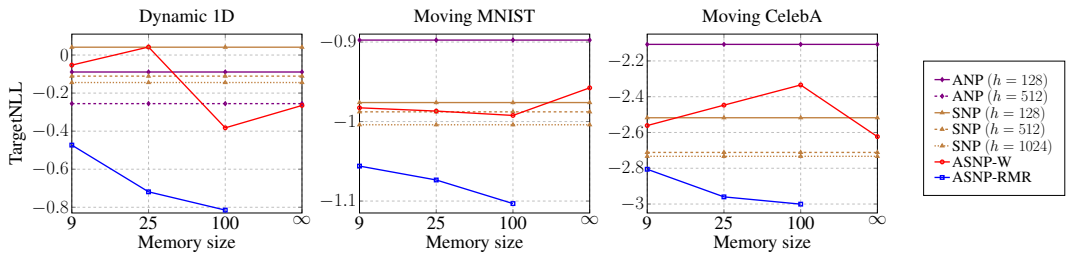
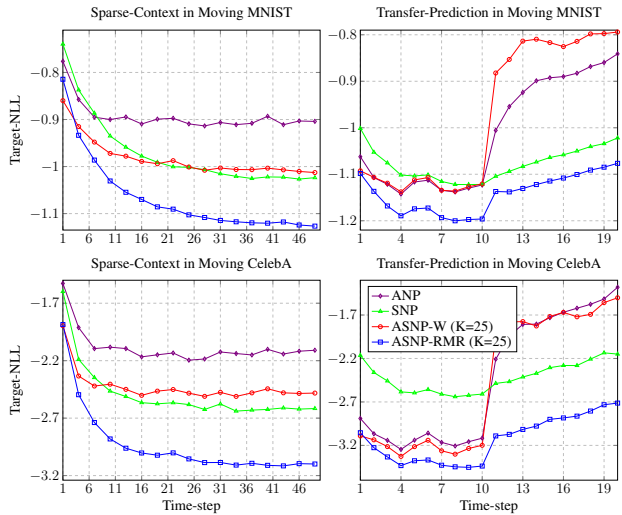
\begin{figure}[t]
    \centering
    \begin{tikzpicture}[scale=0.38]
    \begin{groupplot}[group style={group size=2 by 2,horizontal sep = 40pt,vertical sep = 50pt}]
        \centering
        \nextgroupplot[
            width=9cm,height=7cm,
            scale only axis=true,
            title=\Large{Sparse-Context in Moving MNIST},
            ylabel={Target-NLL},
            xmin=1, xmax=50,
            ymin=-1.135, ymax=-0.71,
            xtick={1,6,11,16,21,26,31,36,41,46},
            ytick={-1.2,-1.1,-1.0,-0.9,-0.8,-0.7,-0.6,-0.5,-0.4},
            xmajorgrids=true,
            ymajorgrids=true,
            legend pos=north east,
            grid style=dashed,
            legend cell align={left},
            label style={font=\Large},
            tick label style={font=\Large}  
        ]
        % ANP(h=128)
        \addplot[
            color=violet,
            mark=diamond,
            line width=1pt,
            mark size=2pt,
            ]
            coordinates {(1,-0.7763337886333466)(4,-0.8575434553623199)(7,-0.8951056492328644)(10,-0.9000209480524063)(13,-0.8945965874195099)(16,-0.9094785743951798)(19,-0.8991864892840385)(22,-0.8974649429321289)(25,-0.9092439675331115)(28,-0.9136675190925598)(31,-0.9063466703891754)(34,-0.9108930724859238)(37,-0.9080524727702141)(40,-0.8931299167871475)(43,-0.9110328459739685)(46,-0.9031675732135773)(49,-0.9040072643756867)
            };
        % SNP(h=128)
        \addplot[
            color=green,
            mark=triangle,
            line width=1pt,
            mark size=2pt,
            ]
            coordinates {(1,-0.7397073543071747)(4,-0.8371651774644852)(7,-0.8865452241897583)(10,-0.9350568455457687)(13,-0.9588660019636154)(16,-0.9776655650138855)(19,-0.9908681231737136)(22,-1.000475031733513)(25,-1.0017398023605346)(28,-1.0065981650352478)(31,-1.0152367842197418)(34,-1.0207558172941207)(37,-1.0254205721616745)(40,-1.0220219504833221)(43,-1.0225670874118804)(46,-1.026567976474762)(49,-1.0236844217777252)
            };
        % SNPK(h=128,K=25)
        \addplot[
            color=red,
            mark=o,
            line width=1pt,
            mark size=2pt,
            ]
            coordinates {(1,-0.859845512509346)(4,-0.9150312906503677)(7,-0.9479892712831497)(10,-0.9720624172687531)(13,-0.9779487031698227)(16,-0.9890808832645416)(19,-0.9940048259496689)(22,-0.9874471533298492)(25,-1.0010105019807816)(28,-1.0078998738527298)(31,-1.003091688156128)(34,-1.0062725007534028)(37,-1.0066301745176316)(40,-1.0034521996974946)(43,-1.0070288133621217)(46,-1.0103774392604827)(49,-1.0127523577213287)
            };
        % ASNP(h=128,K=25)
        \addplot[
            color=blue,
            mark=square,
            line width=1pt,
            mark size=2pt,
            ]
            coordinates {(1,-0.8144472694396973)(4,-0.9335899442434311)(7,-0.9862386095523834)(10,-1.0306332731246948)(13,-1.0546937561035157)(16,-1.0699018508195877)(19,-1.0856676995754242)(22,-1.0905699330568313)(25,-1.1026734614372253)(28,-1.1083219349384308)(31,-1.1145482921600343)(34,-1.1174258732795714)(37,-1.1196552586555482)(40,-1.1204173409938811)(43,-1.1177981305122375)(46,-1.1242400789260865)(49,-1.126620236635208)
            };
        \nextgroupplot[
            width=9cm,height=7cm,
            scale only axis=true,
            title=\Large{Transfer-Prediction in Moving MNIST},
            xmin=1, xmax=20,
            ymin=-1.22, ymax=-0.79,
            xtick={1,4,7,10,13,16,19},
            ytick={-1.2,-1.1,-1.0,-0.9,-0.8,-0.7,-0.6,0.2,0.5},
            xmajorgrids=true,
            ymajorgrids=true,
            grid style=dashed,
            legend cell align={left},
            label style={font=\Large},
            tick label style={font=\Large}  
        ]
        % ANP(h=128)
        \addplot[
            color=violet,
            mark=diamond,
            line width=1pt,
            mark size=2pt,
            ]
            coordinates {(1,-1.0623034465312957)(2,-1.1070595270395278)(3,-1.12156154692173)(4,-1.1426147049665452)(5,-1.1166182160377502)(6,-1.1128233641386032)(7,-1.1345474100112916)(8,-1.138063730597496)(9,-1.12982102394104)(10,-1.1230295884609223)(11,-1.0056077295541763)(12,-0.9545234853029251)(13,-0.9240431386232376)(14,-0.8989572304487229)(15,-0.8924733299016953)(16,-0.8899452841281891)(17,-0.8827234590053559)(18,-0.8683933186531066)(19,-0.859731388092041)(20,-0.8409995698928833)
            };
        % SNP(h=128)
        \addplot[
            color=green,
            mark=triangle,
            line width=1pt,
            mark size=2pt,
            ]
            coordinates {(1,-1.00138398706913)(2,-1.0526082557439804)(3,-1.0755699986219407)(4,-1.1017055082321168)(5,-1.1036187463998794)(6,-1.101377067565918)(7,-1.1154883372783662)(8,-1.1219534289836883)(9,-1.122173103094101)(10,-1.1218612849712373)(11,-1.1043362414836884)(12,-1.0938389736413956)(13,-1.083083667755127)(14,-1.073641867041588)(15,-1.0639572459459306)(16,-1.0581084501743316)(17,-1.050079299211502)(18,-1.0404833626747132)(19,-1.0341145688295363)(20,-1.0218239259719848)
            };
        % SNPK(h=128,K=25)
        \addplot[
            color=red,
            mark=o,
            line width=1pt,
            mark size=2pt,
            ]
            coordinates {(1,-1.0926031267642975)(2,-1.1066607600450515)(3,-1.1192169922590256)(4,-1.1377796918153762)(5,-1.1121954023838043)(6,-1.1070954543352127)(7,-1.1344712764024734)(8,-1.136305359005928)(9,-1.126696479320526)(10,-1.1208783996105194)(11,-0.8822017723321914)(12,-0.8531025540828705)(13,-0.8136216348409653)(14,-0.8095639771223069)(15,-0.816851641535759)(16,-0.825378737449646)(17,-0.8141843438148498)(18,-0.7981517714262009)(19,-0.7975233405828476)(20,-0.7940976268053055)
            };
        % ASNP(h=128,K=25)
        \addplot[
            color=blue,
            mark=square,
            line width=1pt,
            mark size=2pt,
            ]
            coordinates {(1,-1.0982969462871552)(2,-1.1367460072040558)(3,-1.1681902694702149)(4,-1.189403507709503)(5,-1.1743973624706268)(6,-1.1726055139303206)(7,-1.1933035838603974)(8,-1.2000998258590698)(9,-1.1973591291904448)(10,-1.1962266862392426)(11,-1.137113902568817)(12,-1.1379271757602691)(13,-1.1306007993221283)(14,-1.1221206378936768)(15,-1.1145955324172974)(16,-1.1081455194950103)(17,-1.1005472886562346)(18,-1.0910508304834365)(19,-1.0847906845808029)(20,-1.0765328496694564)
            };
        \nextgroupplot[
            width=9cm,height=7cm,
            scale only axis=true,
            title=\Large{Sparse-Context in Moving CelebA},
            xlabel={Time-step},
            ylabel={Target-NLL},
            xmin=1, xmax=50,
            ymin=-3.24, ymax=-1.5,
            xtick={1,6,11,16,21,26,31,36,41,46},
            ytick={-3.2,-2.7,-2.2,-1.7,-1.2},
            xmajorgrids=true,
            ymajorgrids=true,
            legend pos=north east,
            grid style=dashed,
            legend cell align={left},
            label style={font=\Large},
            tick label style={font=\Large}  
        ]
        % ANP(h=128)
        \addplot[
            color=violet,
            mark=diamond,
            line width=1pt,
            mark size=2pt,
            ]
            coordinates {(1,-1.527778001241386)(4,-1.9121553659439088)(7,-2.094967131614685)(10,-2.0825439842045306)(13,-2.0944729460775853)(16,-2.1667237305641174)(19,-2.148835878074169)(22,-2.133098458945751)(25,-2.1955708396434783)(28,-2.1851807332038877)(31,-2.1240775975957513)(34,-2.1378454130142925)(37,-2.1500308947265148)(40,-2.1011965469270946)(43,-2.142068771123886)(46,-2.117610468715429)(49,-2.109272560477257)
            };
        % SNP(h=128)
        \addplot[
            color=green,
            mark=triangle,
            line width=1pt,
            mark size=2pt,
            ]
            coordinates {(1,-1.5982754963636399)(4,-2.1892272877693175)(7,-2.3472091364860534)(10,-2.46484188914299)(13,-2.5119236308336257)(16,-2.565512019395828)(19,-2.5752838110923766)(22,-2.566184149980545)(25,-2.582149600982666)(28,-2.6246002435684206)(31,-2.576558555364609)(34,-2.6388269686698913)(37,-2.6301217210292815)(40,-2.6243009757995606)(43,-2.611080757379532)(46,-2.6212499070167543)(49,-2.6165427803993224)
            };
        % SNPK(h=128,K=25)
        \addplot[
            color=red,
            mark=o,
            line width=1pt,
            mark size=2pt,
            ]
            coordinates {(1,-1.8953582119196652)(4,-2.333564379811287)(7,-2.420707221031189)(10,-2.404629989862442)(13,-2.4496514987945557)(16,-2.5016404259204865)(19,-2.4648998129367827)(22,-2.450974336862564)(25,-2.482948125600815)(28,-2.5104253089427946)(31,-2.475462191104889)(34,-2.5109466671943665)(37,-2.4801814162731173)(40,-2.4447463846206663)(43,-2.4797855520248415)(46,-2.4850269496440887)(49,-2.481989401578903)
            };
        % ASNP(h=128,K=25)
        \addplot[
            color=blue,
            mark=square,
            line width=1pt,
            mark size=2pt,
            ]
            coordinates {(1,-1.885983733162284)(4,-2.4961390709877014)(7,-2.738314137458801)(10,-2.8811318707466125)(13,-2.961755654811859)(16,-3.00319504737854)(19,-3.022657825946808)(22,-3.002336564064026)(25,-3.054815955162048)(28,-3.08689701795578)(31,-3.0860259461402895)(34,-3.108681604862213)(37,-3.0929628801345825)(40,-3.1108091354370115)(43,-3.1156214451789856)(46,-3.096177761554718)(49,-3.0992203068733217)
            };
        \nextgroupplot[
            width=9cm,height=7cm,
            scale only axis=true,
            title=\Large{Transfer-Prediction in Moving CelebA},
            legend style={font=\Large},
            legend pos=north west,
            xlabel={Time-step},
            xmin=1, xmax=20,
            ymin=-3.6, ymax=-1.3,
            xtick={1,4,7,10,13,16,19},
            ytick={-3.7,-3.2,-2.7,-2.2,-1.7,-1.2,-0.7},
            xmajorgrids=true,
            ymajorgrids=true,
            grid style=dashed,
            legend cell align={left},
            label style={font=\Large},
            tick label style={font=\Large}  
        ]
        % ANP(h=128)
        \addplot[
            color=violet,
            mark=diamond,
            line width=1pt,
            mark size=2pt,
            ]
            coordinates {(1,-2.8901365634799006)(2,-3.0659131205081938)(3,-3.142213662862778)(4,-3.2468083667755128)(5,-3.139740219116211)(6,-3.058714543581009)(7,-3.165301411151886)(8,-3.2065048813819885)(9,-3.15746328830719)(10,-3.1169151401519777)(11,-2.2079522931575775)(12,-1.9405538475513457)(13,-1.8078179275989532)(14,-1.804877028465271)(15,-1.7302932846546173)(16,-1.6684854745864868)(17,-1.6220165032148361)(18,-1.5768806472420693)(19,-1.5159268222004174)(20,-1.380546294823289)
            };
        % SNP(h=128)
        \addplot[
            color=green,
            mark=triangle,
            line width=1pt,
            mark size=2pt,
            ]
            coordinates {(1,-2.1641844913363455)(2,-2.3613989049196245)(3,-2.4584549668431284)(4,-2.5841858196258545)(5,-2.595728496313095)(6,-2.557299792766571)(7,-2.6101329827308657)(8,-2.6400325989723203)(9,-2.623856992721558)(10,-2.609354852437973)(11,-2.486743873357773)(12,-2.4666367983818054)(13,-2.412908161878586)(14,-2.370414469242096)(15,-2.30541324198246)(16,-2.280753959417343)(17,-2.2812395083904264)(18,-2.2072109937667848)(19,-2.13523698925972)(20,-2.14942747592926)
            };
        % SNPK(h=128,K=25)
        \addplot[
            color=red,
            mark=o,
            line width=1pt,
            mark size=2pt,
            ]
            coordinates {(1,-3.0910552299022673)(2,-3.13590295791626)(3,-3.2135798048973085)(4,-3.3266061782836913)(5,-3.214173357486725)(6,-3.1409802842140198)(7,-3.2621745586395265)(8,-3.3009407925605774)(9,-3.2352052080631255)(10,-3.198627678155899)(11,-1.9363369911909103)(12,-1.7803323292732238)(13,-1.7755840834975243)(14,-1.8260747480392456)(15,-1.7173162412643432)(16,-1.669397594332695)(17,-1.7206568658351897)(18,-1.6925308281183242)(19,-1.557214994430542)(20,-1.5016947188973426)
            };
        % ASNP(h=128,K=25)
        \addplot[
            color=blue,
            mark=square,
            line width=1pt,
            mark size=2pt,
            ]
            coordinates {(1,-3.054080775976181)(2,-3.2262208962440493)(3,-3.332598807811737)(4,-3.4321782994270325)(5,-3.376622359752655)(6,-3.367646849155426)(7,-3.42778968334198)(8,-3.446378264427185)(9,-3.4522285795211793)(10,-3.4376662588119506)(11,-3.090925419330597)(12,-3.0729705142974852)(13,-3.016162815093994)(14,-2.9777334570884704)(15,-2.9000813341140748)(16,-2.8839481377601626)(17,-2.863220888376236)(18,-2.804712386131287)(19,-2.732153193950653)(20,-2.713148807287216)
            };
        \legend{ANP,SNP,ASNP-W (K=25),ASNP-RMR (K=25)}
    \end{groupplot}
    \end{tikzpicture}
    \vspace{-1mm}
\caption{Target-NLL computed at each time-step in moving CelebA image completion task sequence. Hidden unit size in all models is $128$.}
\label{fig:2d_reg_fig}
\vspace{-4mm}
\end{figure}

\textbf{Target-NLL.} We plot the target-NLLs for each task-step in Fig.~\ref{fig:1d_reg_graph}. Consider the performance of ASNP-W to study the effect of extending SNP with the standard attention. We observe that improvements compared to SNP are small. Comparing these with ASNP-RMR shows that sequential latent and the standard attention alone cannot solve underfitting without addressing sparse and obsolete contexts. In \textit{transfer-prediction} regime, for $t\leq10$, we note that target-NLLs of ANP are similar to that of ASNP-RMR. This is because, in this interval, ANP exploits the larger context sizes without attending to the obsolete points. However, the main focus of this setting is multi-step transfer for $t>10$ on which ANP degrades quickly.

\textbf{Effect of Memory Size and Latent Size.} In Fig.~\ref{fig:k_fig}, we show the effect of memory and latent sizes on performance. We observe that increasing the latent size in SNP from $128 \text{ to } 1024$ shows small improvements with diminishing returns. This shows that a larger latent size is not adequate. Additionally, we observe that ASNP-W does not surpass ANP significantly. Hence, we conclude that the recurrent latents $z_{\leq t}$ fail to inform the observation model about the shift in the obsolete points. We further note that ASNP-RMR at $K=9$ outperforms ASNP-W for all choices of $K$. This shows that RMR is more size-efficient.

\textbf{Training Time.} In Fig.~\ref{fig:1d_reg_graph}, we plot the target-NLL against the training wall-clock time, and we note that ASNP-RMR imposes no significant overhead in convergence. 

\textbf{Qualitative Analysis.} In Fig.~\ref{fig:1d_reg_sam}, we show the predictive means of the target function. We observe that SNP underfits the target function in multiple locations. Furthermore, in ANP and ASNP-W, attention helps the prediction only at those obsolete points that did not undergo a significant shift. In these models, attention also causes misestimation of the function when obsolete points shift significantly. It shows that without addressing the obsolete context issue, the standard attention cannot resolve underfitting. In comparison, ASNP-RMR makes more accurate predictions.

\subsection{Dynamic 2D Image Completion}

In this setting, the task is to complete an image by estimating the pixel value at a given pixel location. The image belongs to a sequence of images that contains a moving image patch on a white canvas. In this dynamic setting, the model must not only estimate the unseen pixel values but also track their motion. The moving images are taken from the MNIST~\citep{lecun1998gradient} and the CelebA~\citep{liu2015faceattributes} datasets, and hence we call these settings \textit{moving MNIST} and \textit{moving CelebA}, respectively.

\textbf{Target NLL.} We plot the target-NLLs in Fig.~\ref{fig:2d_reg_fig}. Recall that in the sparse-context regime, a model must transfer-learn at every task-step, and in the transfer-prediction regime, a model must transfer-learn on task-steps $t>10$ that have an empty context. We observe that when such transfer-learning is required, the performance of ASNP-W degrades compared to SNP. This implies that attention on obsolete points is not only ineffective but also detrimental.

\textbf{Effect of Memory and Latent Sizes.} From Fig.~\ref{fig:k_fig}, our conclusions about the effect of memory size and latent size are similar to those in dynamic 1D regression. ASNP-RMR is not only size-efficient but also shows monotonic improvement as we increase the memory size. Note that in Fig.~\ref{fig:k_fig}, we do not test ANP using a larger latent size because we find its memory usage too large to train.

\begin{figure}[t]
        \centering
        \includegraphics[width=1.0\linewidth]{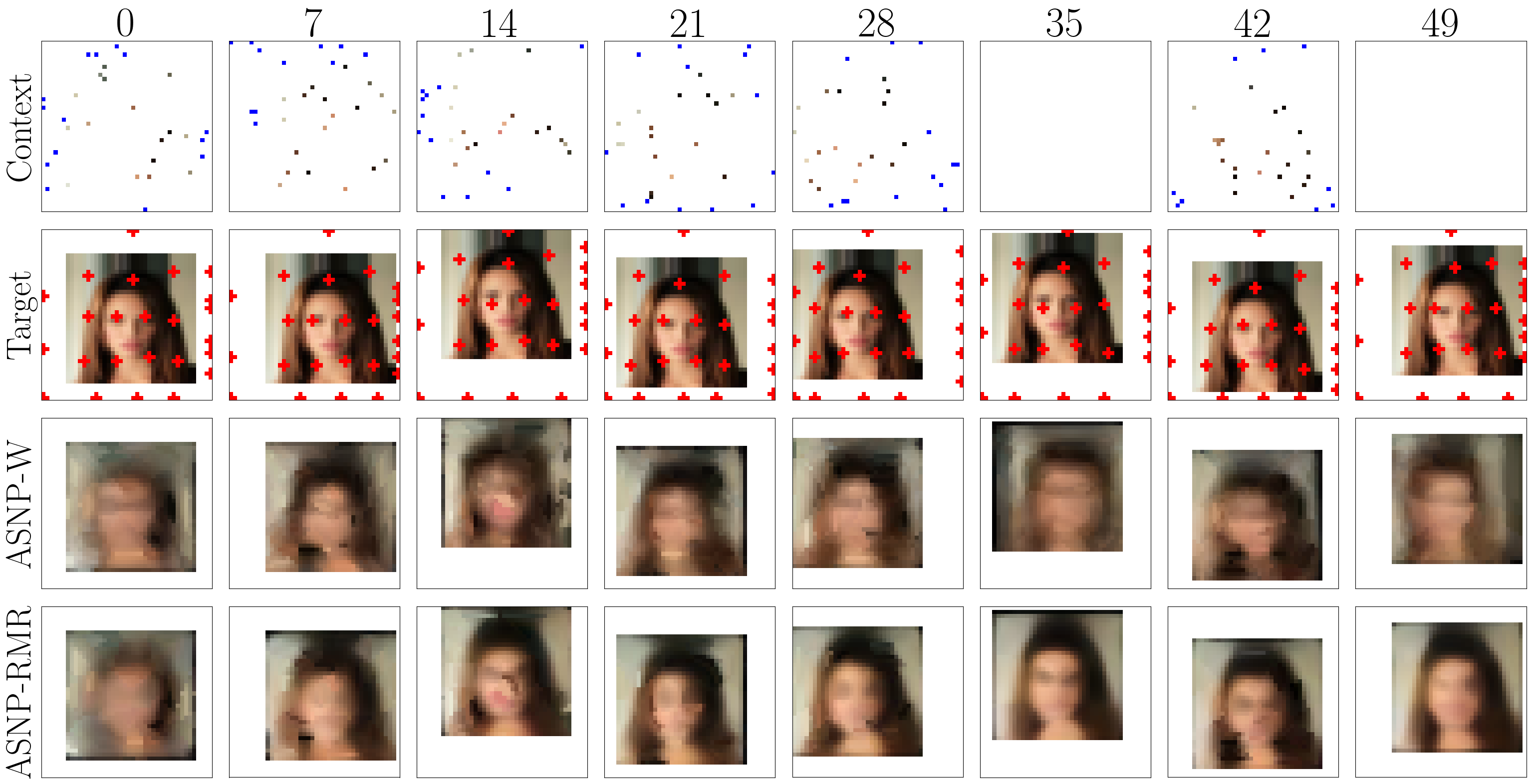}
    \vspace{-5mm}
    \caption{Qualitative evaluation on moving CelebA image completion task sequence. \textit{Red dots:} Imaginary keys imagined by ASNP-RMR. The memory size $K$ is $25$.}
    \label{fig:im_moving_fig}
    \vspace{-3mm}
\end{figure}

\begin{figure}[t]
        \centering
        \includegraphics[width=1.0\linewidth]{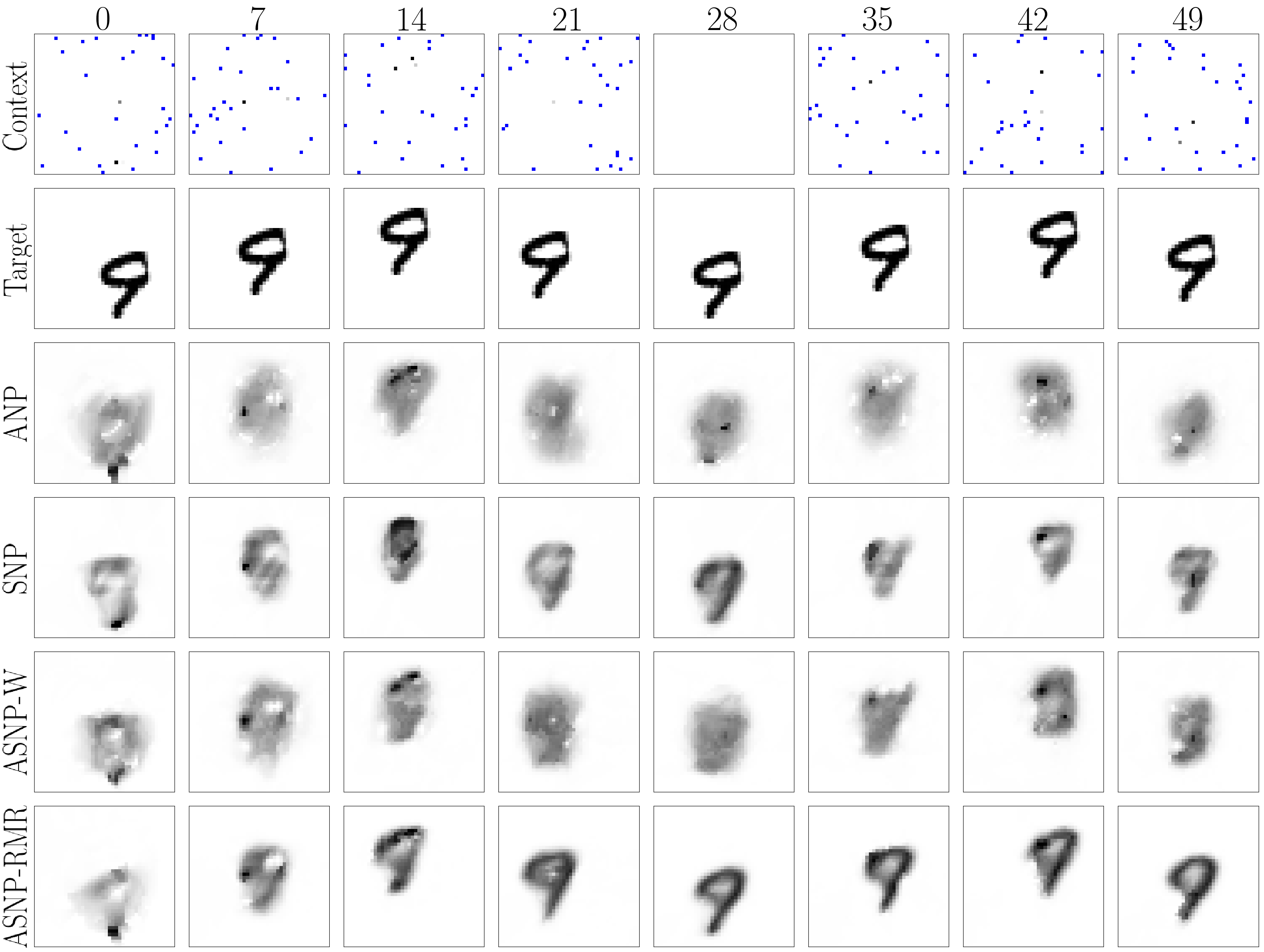}
    \vspace{-3mm}
    \captionof{figure}{Moving MNIST image completion samples in the sparse-context regime. Memory size of ASNP-RMR is $25$.}
    \vspace{-3mm}
    \label{fig:2d_reg_sam}
    \vspace{0mm}
\end{figure}

\textbf{Qualitative Analysis.}
Fig.~\ref{fig:im_moving_fig} shows the positions of imaginary keys generated by RMR. We observe that the model learns to discover certain key points on the CelebA patch, and it tracks them as the patch moves. It also places some keys on the edges of the canvas to account for the bounces. Fig.~\ref{fig:2d_reg_sam} shows qualitative samples for moving MNIST image completion in sparse-context setting. As ASNP-RMR accumulates more context, RMR reforms the obsolete points for the current task. Consequently, we observe that the predicted images become clearer over time. We also observe that in comparison to SNP, ASNP-W shows deterioration. In Fig.~\ref{fig:2d_reg_sam} and~\ref{fig:im_moving_fig}, consider the task-steps with an empty context. On these task-steps, ANP and ASNP-W predict poor quality images compared to ASNP-RMR, which shows that the standard attention cannot deal with the obsolete context.

\subsection{Ablation Study on RMR}
\begin{figure}
    \centering
    \begin{tikzpicture}[scale=0.38]
    \begin{groupplot}[group style={group size=2 by 2,horizontal sep = 40pt,vertical sep = 50pt}]
        \centering
        \nextgroupplot[
            width=9cm,height=7cm,
            scale only axis=true,
            title=\Large{Sparse-Context in 1D GP Regression},
            ylabel={Target-NLL},
            xmin=1, xmax=50,
            ymin=-1.23, ymax=2.1,
            xtick={1,6,11,16,21,26,31,36,41,46},
            ytick={-1.0,-0.5,0.0,0.5,1.0,1.5,2.0},
            xmajorgrids=true,
            ymajorgrids=true,
            legend pos=north east,
            grid style=dashed,
            label style={font=\Large},
            tick label style={font=\Large}  
        ]
        % ASNPw/oFT
        \addplot[
            color=magenta,
            mark=square,
            line width=1pt,
            mark size=2pt,
            ]
            coordinates {
            (1,1.9269291079044342)(4,1.0880254736542703)(7,0.5433863772032782)(10,0.27150505703408273)(13,0.10386992930434644)(16,-0.04835733529587742)(19,-0.12051325572305359)(22,-0.21016233979258686)(25,-0.28276083814445885)(28,-0.35487621252657847)(31,-0.39293895691633224)(34,-0.45248841217719016)(37,-0.49361345544457436)(40,-0.48479209241457283)(43,-0.45835863282904027)(46,-0.4264550095051527)(49,-0.39884001556783916)
            };
        % ASNPw/oAtt
        \addplot[
            color=brown,
            mark=square,
            line width=1pt,
            mark size=2pt,
            ]
            coordinates {
            (1,1.9079117619991302)(4,0.9581248050928116)(7,0.39018968494608997)(10,0.03773805968929082)(13,-0.21815182768332306)(16,-0.42782094607129695)(19,-0.5840360134840011)(22,-0.6952031992375851)(25,-0.74661875218153)(28,-0.7927564844489098)(31,-0.8339820328354836)(34,-0.8911135423183442)(37,-0.9095165717601776)(40,-0.8943898928165436)(43,-0.8605753186345101)(46,-0.8404517608880997)(49,-0.8165441164374352)
            };
        % ASNP
        \addplot[
            color=blue,
            mark=square,
            line width=1pt,
            mark size=2pt,
            ]
            coordinates {(1,1.9129909467697144)(4,0.9779843148589135)(7,0.3168356418143958)(10,-0.09146490456303581)(13,-0.41183125554583966)(16,-0.6737789215054363)(19,-0.8363264921680093)(22,-0.9442536509037018)(25,-1.0093567198514939)(28,-1.0559395164251328)(31,-1.0960114961862564)(34,-1.1337012648582458)(37,-1.1566540223360062)(40,-1.1268552184104919)(43,-1.0795878905057907)(46,-1.0229877680540085)(49,-0.9684694135189056)
            };
        \nextgroupplot[
            width=9cm,height=7cm,
            scale only axis=true,
            title=\Large{Transfer-Prediction in 1D GP regression},
            xmin=1, xmax=20,
            ymin=-1.35, ymax=-0.39,
            xtick={1,4,7,10,13,16,19},
            ytick={-1.3,-1.0,-0.7,-0.4,-0.1,0.2,0.5},
            xmajorgrids=true,
            ymajorgrids=true,
            grid style=dashed,
            label style={font=\Large},
            tick label style={font=\Large}  
        ]
        % ASNPw/oFT
        \addplot[
            color=magenta,
            mark=square,
            line width=1pt,
            mark size=2pt,
            ]
            coordinates {
            (1,-0.7798858299478888)(2,-1.1126454737782479)(3,-1.1754855014570058)(4,-1.2061401450634002)(5,-1.2311547964811325)(6,-1.2528380858898163)(7,-1.2597840571403502)(8,-1.259891264438629)(9,-1.2662813472747803)(10,-1.268920075893402)(11,-1.2586091005802154)(12,-1.2256335484981538)(13,-1.1841600972414017)(14,-1.135165785551071)(15,-1.0858278435468673)(16,-1.0283783406019211)(17,-0.9685910922288895)(18,-0.905061976313591)(19,-0.8366881507635117)(20,-0.7537605220079422)
            };
        % ASNPw/oAtt
        \addplot[
            color=brown,
            mark=square,
            line width=1pt,
            mark size=2pt,
            ]
            coordinates {
            (1,-0.8798379880189896)(2,-0.9650866570323706)(3,-1.0662318041920662)(4,-1.078687037229538)(5,-1.0580044573545455)(6,-1.0861684900522233)(7,-1.1023639652132988)(8,-1.0770682621002197)(9,-1.0152478086948395)(10,-1.101384374499321)(11,-0.7215475258231163)(12,-0.71653787702322)(13,-0.6876495349407196)(14,-0.6515300525724887)(15,-0.6145750043541193)(16,-0.5778340893983841)(17,-0.5412132430076599)(18,-0.5054182265355485)(19,-0.46966241983696816)(20,-0.4338068864587694)
            };
        % ASNP
        \addplot[
            color=blue,
            mark=square,
            line width=1pt,
            mark size=2pt,
            ]
            coordinates {(1,-0.9090237456141039)(2,-1.2101020067930222)(3,-1.2765457516908645)(4,-1.3067767804861068)(5,-1.3242017579078675)(6,-1.3321631848812103)(7,-1.3378321528434753)(8,-1.3395658266544341)(9,-1.3391781103610993)(10,-1.3401928889751433)(11,-1.3356824505329132)(12,-1.3333749818801879)(13,-1.3274410355091095)(14,-1.3204669165611267)(15,-1.3118839299678802)(16,-1.3004737424850463)(17,-1.2869476616382598)(18,-1.2719783198833465)(19,-1.2554468202590943)(20,-1.2380664789676665)
            };
        \nextgroupplot[
            width=9cm,height=7cm,
            scale only axis=true,
            title=\Large{Sparse-Context in 2D Moving MNIST},
            xlabel={Time-step},
            ylabel={TargetNLL},
            xmin=1, xmax=50,
            ymin=-1.15, ymax=-0.79,
            xtick={1,6,11,16,21,26,31,36,41,46},
            ytick={-1.1,-1.05,-1.0,-0.95,-0.9,-0.85,-0.8,-0.75,-0.7,-0.65,-0.6},
            xmajorgrids=true,
            ymajorgrids=true,
            legend pos=north east,
            grid style=dashed,
            label style={font=\Large},
            tick label style={font=\Large}  
        ]
        % ASNPw/oFT
        \addplot[
            color=magenta,
            mark=square,
            line width=1pt,
            mark size=2pt,
            ]
            coordinates {
            (1,-0.8673984467983246)(4,-0.9260965824127197)(7,-0.9610925149917603)(10,-0.966975821852684)(13,-0.9755453383922577)(16,-0.9845874303579331)(19,-0.9871069699525833)(22,-0.9863150852918625)(25,-0.9889589893817902)(28,-0.9919773346185684)(31,-0.9923447597026825)(34,-0.9966348695755005)(37,-0.9982608902454376)(40,-0.9955677366256714)(43,-0.9944038689136505)(46,-0.9968101143836975)(49,-0.9977513533830643)
            };
        % ASNPw/oAtt
        \addplot[
            color=brown,
            mark=square,
            line width=1pt,
            mark size=2pt,
            ]
            coordinates {
            (1,-0.8511062443256379)(4,-0.9223429501056671)(7,-0.94334368288517)(10,-0.9612269419431686)(13,-0.9721283048391343)(16,-0.9813009619712829)(19,-0.9902136236429214)(22,-0.9931610453128815)(25,-1.001476407647133)(28,-1.0059425932168962)(31,-1.007022699713707)(34,-0.9991022092103958)(37,-1.0033267945051194)(40,-1.0111114907264709)(43,-1.0110860645771027)(46,-1.0078330439329148)(49,-1.0094502568244934)
            };
        % ASNP
        \addplot[
            color=blue,
            mark=square,
            line width=1pt,
            mark size=2pt,
            ]
            coordinates {(1,-0.8144472694396973)(4,-0.9335899442434311)(7,-0.9862386095523834)(10,-1.0306332731246948)(13,-1.0546937561035157)(16,-1.0699018508195877)(19,-1.0856676995754242)(22,-1.0905699330568313)(25,-1.1026734614372253)(28,-1.1083219349384308)(31,-1.1145482921600343)(34,-1.1174258732795714)(37,-1.1196552586555482)(40,-1.1204173409938811)(43,-1.1177981305122375)(46,-1.1242400789260865)(49,-1.126620236635208)
            };
        \nextgroupplot[
            width=9cm,height=7cm,
            scale only axis=true,
            title=\Large{Transfer-Prediction in 2D Moving MNIST},
            xlabel={Time-step},
            xmin=1, xmax=20,
            ymin=-1.22, ymax=-0.94,
            xtick={1,4,7,10,13,16,19},
            ytick={-1.2,-1.1,-1.0,-0.9,-0.8,-0.7},
            xmajorgrids=true,
            ymajorgrids=true,
            grid style=dashed,
            legend style={font=\Large},
            label style={font=\Large},
            tick label style={font=\Large}  
        ]
        % ASNPw/oFT
        \addplot[
            color=magenta,
            mark=square,
            line width=1pt,
            mark size=2pt,
            ]
            coordinates {(1,-1.1110208296775819)(2,-1.1689755809307099)(3,-1.1672075307369232)(4,-1.1755707776546478)(5,-1.1734219974279403)(6,-1.178192989230156)(7,-1.1879086136817931)(8,-1.18235125541687)(9,-1.186472043991089)(10,-1.1902238309383393)(11,-1.1312517023086548)(12,-1.111848196387291)(13,-1.0940753936767578)(14,-1.0806309777498244)(15,-1.0677478468418122)(16,-1.0600218737125398)(17,-1.0469635343551635)(18,-1.03695598423481)(19,-1.0249821090698241)(20,-1.0142359948158264)
            };
        % ASNPw/oAtt
        \addplot[
            color=brown,
            mark=square,
            line width=1pt,
            mark size=2pt,
            ]
            coordinates {
            (1,-1.1092426776885986)(2,-1.1238842052221298)(3,-1.1363699376583098)(4,-1.1591873580217362)(5,-1.1316184604167938)(6,-1.12930144906044)(7,-1.1531272900104523)(8,-1.1568533486127854)(9,-1.149526015520096)(10,-1.1482304102182388)(11,-1.0274429678916932)(12,-1.01991901576519)(13,-1.0094276010990142)(14,-1.003597201704979)(15,-0.9962067544460297)(16,-0.987973068356514)(17,-0.9772955924272537)(18,-0.9691407835483551)(19,-0.9625348055362701)(20,-0.9510930222272873)
            };
        % ASNP
        \addplot[
            color=blue,
            mark=square,
            line width=1pt,
            mark size=2pt,
            ]
            coordinates {(1,-1.0982969462871552)(2,-1.1367460072040558)(3,-1.1681902694702149)(4,-1.189403507709503)(5,-1.1743973624706268)(6,-1.1726055139303206)(7,-1.1933035838603974)(8,-1.2000998258590698)(9,-1.1973591291904448)(10,-1.1962266862392426)(11,-1.137113902568817)(12,-1.1379271757602691)(13,-1.1306007993221283)(14,-1.1221206378936768)(15,-1.1145955324172974)(16,-1.1081455194950103)(17,-1.1005472886562346)(18,-1.0910508304834365)(19,-1.0847906845808029)(20,-1.0765328496694564)
            };
        \legend{ASNP-RMR (no flow-tracking), ASNP-RMR (no flow-interaction), ASNP-RMR}
     \end{groupplot}   
    \end{tikzpicture}
    \vspace{-5mm}
\caption{Target-NLL for ASNP-RMR compared against two alternate versions by way of ablation -- \textit{i)} without flow-tracking and \textit{ii)} without flow-interaction.}
\label{fig:abla_fig}
    \vspace{-3mm}
\end{figure}
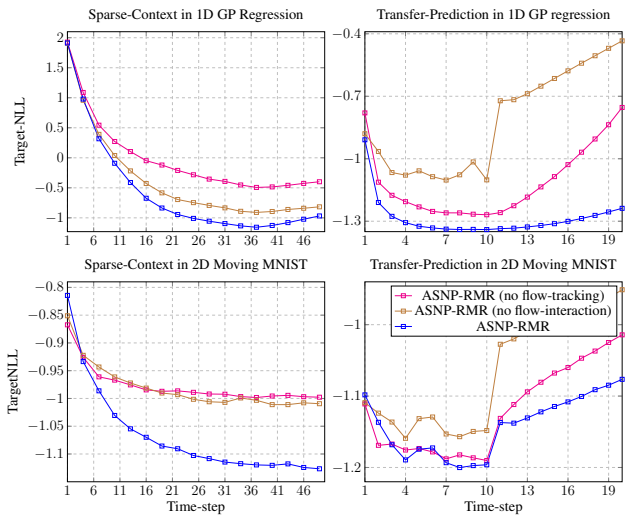

In this section, we perform an ablation on RMR to create two modifications, and we report the results in Fig.~\ref{fig:abla_fig}. 

\textit{i) No Value-Flow Tracking.} In this modification, the model generates new imaginary values only via self-attention on the previous values. Thus, without the recurrence, we expect the model to forget the transition dynamics. In Fig.~\ref{fig:abla_fig}, we observe in the transfer-prediction setting that the performance deteriorates sharply in task-steps $[11,20]$. This is because the context is empty, and value imagination fails to extrapolate without capturing the transition dynamics. 

\textit{ii) No Value-Flow Interaction.} In this modification, the model generates new imaginary values only via Value-Flow Tracking. To incorporate the real context, we provide it as an input to the flow tracking RNNs. Because of no value interaction, we observe performance degradation in Fig.~\ref{fig:abla_fig}. Thus, we conclude that to perform effective value imagination, the model requires both flow tracking and interaction.

\subsection{Dynamic 2D Image Rendering}
In this setting, we consider a sequence of images as in the moving CelebA dataset. The task for the model is to take a location on the canvas and predict the image-patch centered at the location. Because the task is to generate an image, we replace the SNP baseline with a TGQN \cite{snp}. Similarly, we replace ASNP-RMR and ASNP-W with ATGQN-RMR and ATGQN-W, respectively. Here, ATGQN stands for attentive TGQN -- a model that we develop which incorporates attention into TGQN. Hence, ATGQN-RMR is TGQN equipped with attention on RMR, and ATGQN-W is equipped with the standard attention on a context window. See Appendix \ref{app:rendering_details} for implementation details. We report the results on these models in Table~\ref{tab:atgqn_tab}, and we note that ATGQN-RMR outperforms ATGQN-W and TGQN.

\begin{table}[t]
\footnotesize
\centering
\begin{tabular}{p{2.9cm}p{0.9cm}p{0.9cm}p{0.8cm}}
\toprule
Regime                & ATGQN -RMR   & ATGQN -W & TGQN \\ \midrule
\footnotesize Sparse-Context & \textbf{2.57}& 2.77 & 3.23  \\
\footnotesize Transfer-Prediction   & \textbf{2.04}& 2.8 & 3.08 \\ \bottomrule
\end{tabular}
\caption{Target-NLL comparison between ATGQN-RMR, ATGQN-W and TGQN in moving CelebA image rendering task.}\label{tab:atgqn_tab}
\vspace{0mm}
\end{table}

\section{Conclusion}
In this paper, we argued that the two problems, sparse context and obsolete context, observed in meta-transfer learning due to temporal task-shift, make the underfitting issue in SNP more severe. Then, to resolve this, we proposed a novel attention model using imaginary context generated by Recurrent Memory Reconstruction (RMR) and a robust probabilistic meta-transfer learning model, Attentive Sequential Neural Processes. Our experiments demonstrate that existing methods show weaknesses in dealing with sparse and obsolete context, and that using RMR-based attention in SNP is an effective way to resolve the issue. The ablation study shows that the recurrent context modeling and the interaction model are the key components to achieve this improved robustness. In the future, it will be interesting to apply this robust model to various other problems including meta-transfer reinforcement learning.
% \newpage 
\section*{Acknowledgements}
JY thanks Kakao Brain and SAP for their support and Hoyong Jang for helpful feedback. SA thanks Kakao Brain and Center for Super Intelligence (CSI) for their support. 

\bibliography{refs}
\bibliographystyle{icml2020}

\newpage
\clearpage

% comment appendix part
\begin{appendices}

\onecolumn
\section{Model Details} \label{app:model_details}
\subsection{ANP, SNP and ASNPs}

In this section, we describe the model details of ANP, SNP, ASNP-W, and ASNP-RMR. All the models share the basic components such as the real context encoder and decoder. There are two context encoders for producing the latent and the deterministic representations. \textit{i) Latent Encoder:} In this encoder, we first use a 3-layer MLP with ReLU~\citep{nair2010rectified} activation to encode each context point. Next, we sum-pool these encoded representations to obtain a single context representation. Lastly, we provide this representation to a 2-layer MLP to compute the mean and variance of the latent. \textit{ii) Deterministic Encoder:} In this encoder, we use a 6-layer MLP with ReLU activation to encode each context point. For SNP, we sum-pool these encodings. For attention-based models, we associate each encoding with its corresponding key to obtain a key-value set. For ANP, we directly attend to this key-value set based on the target queries. For ASNP-W and ASNP-RMR, we first augment this key-value set with a memory (i.e.~RMR or context buffer) and then attend on it given the target queries.

\textbf{Attention Mechanism.} In the attention module, ANP, ASNP-W and ASNP-RMR can use any of the following implementations: Dot-Product, Laplace or Multihead. In our experiments, we use Multihead attention~\citep{vaswani2017attention} as recommended in \citet{anp}.
% since it showed the best performance for the Gaussian Process task~\citet{anp}.
ASNP-RMR has two attention modules -- one for value-flow interaction and other for the deterministic encoder. In our implementation, both attention modules share parameters since it showed better performance compared to using different parameters.

\textbf{Encoding task-step into the query.} Recall that in ANP, we perform sum-pooling of context points from different tasks. Similarly, in ASNP-W, we store the context points from different tasks in the same memory buffer. In these situations, if the task-step information is absent from the context points, the models cannot distinguish the task to which each point belongs. To prevent this, we concatenate each query with the corresponding task-step information. The task-step information is provided as a normalized float value $e_t=0.25+0.5(t/T)$, where $t$ is the task-step and $T$ is the length of the task-sequence. Therefore, the augmented query is given by $x'=(x,e_t)$. 

\textbf{Recurrent Modules.} The sequential models~(SNP and ASNP) have an underlying recurrent architecture in the latent encoder and the deterministic encoder. As in SNP \cite{snp}, we use the recurrent state-space model (RSSM)~\citep{hafner2018learning} and we use LSTM to implement it. For RMR, we also use LSTM to implement the recurrent modules in the key and value update. The LSTM module for the key update takes the concatenated keys as an input. We use the default setting of Tensorflow~\citep{abadi2016tensorflow} for all LSTMs. The LSTM hidden unit size and representation size are the same and are denoted by $h$. We experiment with values of $h=$ $128$, $512$ and $1024$.

\textbf{Memory Sizes.} Let $K$ denote the RMR memory size. Let $K$ also denote the size of the context window in ASNP-W. In our experiments, we evaluate the models for $K=$ $9$, $25$ and $100$. We also evaluate the case when ASNP-W stores the entire past context. We denote this case with $K=\mathrm{inf}$. 

\textbf{Training.} The initial imaginary context points are trainable parameters. For dynamic 1D regression, we train our models with a learning rate $10^{-4}$ and batch size 16. For moving MNIST and moving CelebA, we train with batch sizes 8 and 4, respectively.

\subsection{TGQN and ATGQN}
In this section, we describe the model details of TGQN, ATGQN-W and ATGQN-RMR. Each of them share the basic architecture that consists of \textit{i)} an encoder to encode the real contexts,  \textit{ii)} the recurrent state-space model to generate the latents and  \textit{iii)} a decoder to generate the target outputs. In the encoder, we use $3$-layer CNN with ReLU activation to encode the context image. We append the query to the last layer to provide the encoder with the query information similar to the Tower Network in~\citet{eslami2018neural}. As in TGQN~\citet{snp}, we sum-pool the representations to obtain a global representation. To obtain the latents $z_{1:T}$, we use Temporal-ConvDRAW~\citep{snp}, which is implemented using a ConvLSTM~\citep{xingjian2015convolutional}. For the deterministic representation also, we use ConvLSTM. Similar to CGQN~\citep{kumar2018consistent}, our decoder renders the image recurrently. 

In ATGQN-W, we append the task-step information to the queries as described above for ANP and ASNP-W. We implement attention in ATGQN same as ANP and ASNP-W. To generate the imaginary keys we use LSTM with the default Pytorch settings~\citep{paszke2017automatic}. To maintain a convolutional architecture, we implement the value flow-tracker using a ConvLSTM. 

In our experiments, the patch-size is $8$ and $h=128$. Number of DRAW steps for Temporal-ConvDRAW is $3$ and the dimension of $z$ is $4$. At each task-step, we take $z_t$ to be the latent obtained at the last iteration of ConvDRAW. The ConvLSTM hidden unit size in latent and deterministic encoding is $40$. In the decoder, the hidden unit size is $32$ and the number of steps for auto-regressive rendering is $2$. The query size is $2$. For ATGQN-W, the query size is $3$ since we concatenate the query and the task-step. We take the imaginary context size and the context window size as $100$. We train the models with a learning rate $10^{-4}$ and batch size $4$.

\section{Setting Details}\label{app:task_details}
\subsection{Dynamic 1D Regression} \label{app:gp_details}
To generate a task-sequence, we first choose the kernel parameters for the Gaussian Process setting randomly at $t=1$. In our experiments, the GPs use a squared-exponential kernel that are characterized by a length-scale $l$ and a kernel-scale $\sigma$. Therefore, at each task-step, the kernel parameters are updated as $l \gets l + \Delta l + \epsilon_{l}$ and $\sigma \gets \sigma + \Delta \sigma + \epsilon_\sigma$ where $\epsilon_l$ and $\epsilon_\sigma$ are a small Gaussian noise. At $t=1$, we randomly choose $\Delta l$ from $[-0.03,0.03]$ and $\Delta \sigma$ from $[-0.05,0.05]$. We sample the noise values at every task-step from a Gaussian distribution $\mathcal{N}(0,0.1)$.  

In the sparse-context regime, we choose the initial values of $l$ from $[0.7,1.2]$ and $\sigma$ from $[1.0,1.6]$. We then randomly choose $45$ task-steps out of $50$ and provide a small-sized context in them. For the remaining tasks, we provide an empty context. Let $n$ denote the number of contexts and $m$ denote the number of targets. At the task-steps for providing a small-sized context, we provide $1$ context, $n=1$. At every task-step, we randomly select $m$ from $[1,11-n]$. In the transfer-prediction regime, we select the initial values of $l$ from $[1.2,1.9]$ and $\sigma$ from $[1.6,3.1]$. We provide a large-sized context in the first $10$ task-steps out of $20$ and an empty context in the remaining. At the task-steps chosen for providing a non-empty context, we randomly choose $n$ from $[5,50]$. At every task-step, we randomly choose $m$ from $[1,51-n]$. 

\subsection{Dynamic 2D Image Completion} \label{app:2d_details}
In this setting, we experiment on the moving MNIST and moving CelebA dataset. For each task-step, we take a white canvas of size $42 \times 42$ that consists of a moving image. The size of the moving MNIST image is $28 \times 28$ and the size of the moving CelebA face image is $32 \times 32$. At $t=1$, the digits or the faces start to move from a random location and head towards a direction chosen randomly. The motion speed is $3$ pixels per task-step. At every task-step, we also add a small Gaussian noise to the transition to introduce stochasticity similar to the dynamic 1D regression setting. When a wall is encountered, the image performs a perfectly elastic bounce. As in the dynamic 1D regression setting, we take the same sequence length and choose the same number of task-steps for providing a non-empty context. In the sparse-context regime, we take $n=30$ for the tasks with non-empty context. We randomly choose $m$ from $[1, 51-n]$ for every task. In the transfer-prediction regime, for the tasks chosen for a non-empty context, we randomly choose $n$ from $[5,500]$. For every task, we randomly choose $m$ from $[1, 501-n]$.
 
\subsection{Dynamic 2D Image Rendering} \label{app:rendering_details}
In this setting, we experiment on the moving CelebA dataset. We take the size of the white canvas as $80 \times 80$ and the size of the moving face image as $64 \times 64$. The image-patch size is $8 \times 8$. The moving image starts to move from a random location in a random direction at a speed of $13$ pixels per task-step. When a wall is encountered, the image performs a perfect bounce. We add a small Gaussian noise on each transition. We take sequences of length $6$. In the sparse-context regime, we provide a small-sized context in $5$ task-steps and an empty context in the remaining. At the task-steps chosen for providing a small-sized context, we provide $100$ contexts i.e.~$n=100$. At every task-step, we randomly choose $m$ from $[1,151-n]$. In the transfer-prediction regime, we provide a large-sized context in the first $3$ task-steps out of $6$ and an empty context in the remaining. At the task-steps chosen for non-empty context, we randomly choose $n$ from $[300,350]$. At every time-step, we randomly choose $m$ from $[1, 351-n]$.

%\newpage
\section{ELBO Derivations} \label{app:elbo_deri}
Although our proposed framework allows a stochastic RMR, for simplicity, we use a deterministic RMR as $P(\tC_t | \tC_\lt, C_{\leq t}) = \delta[\tC_t = \mathrm{RMR}(\tC_\lt,C_{\leq t})]$ where $\delta$ is a Dirac delta function. With this, we can describe the generative process of ASNP as follows. 
\eq{
\vspace{-2mm}
P(Y,Z,\tC|X,C) = \prod_{t=1}^T P(y_t|x_t,z_t,\barC_t)P(z_t|z_\lt,\tC_{\leq t},C_{\leq t}).
}
For this generative process, we derive an ELBO as follows.

\begin{flalign*}
\begin{aligned}
&\log P(Y|X,C)&\\ 
  &= \log \eE_{Q(Z|\tC,C,D)}\left[\frac{P(Y,Z,\tC|X,C)}{Q(Z|\tC,C,D)}\right]&\\
  &= \log \eE_{Q(Z|\tC,C,D)}\left[\prod_{t=1}^T\frac{P(y_t|x_t,z_t,\barC_t)P(z_t|z_\lt,\tC_{\leq t},C_{\leq t})}{Q(z_t|z_\lt,\tC_{\leq t},C,D)}\right]&\\
  &\geq \eE_{Q(Z|\tC,C,D)}\left[\log  \prod_{t=1}^T\frac{P(y_t|x_t,z_t,\barC_t)P(z_t|z_\lt,\tC_{\leq t},C_{\leq t})}{Q(z_t|z_\lt,\tC_{\leq t},C,D)}\right]&\\
  &= \eE_{Q(Z|\tC,C,D)}\sum_{t=1}^T\left[\log\frac{P(y_t|x_t,z_t,\barC_t)P(z_t|z_\lt,\tC_{\leq t},C_{\leq t})}{Q(z_t|z_\lt,\tC_{\leq t},C,D)}\right]&\\
  &= \eE_{Q(Z|\tC,C,D)}\sum_{t=1}^T\left[\log P(y_t|x_t,z_t,\barC_t) - \log \frac{Q(z_t|z_\lt,\tC_{\leq t},C,D)}{P(z_t|z_\lt,\tC_{\leq t},C_{\leq t})}\right]&\\
  &= \sum_{t=1}^T\eE_{\pd{t'}{t}Q(z_{t'}|z_{<t'},\tC_{\leq t'},C,D)}\left[\log P(y_t|x_t,z_t,\barC_t)\right]&\\
  &\quad- \eE_{\pd{t'}{t}Q(z_{t'}|z_{<t'},\tC_{\leq t'},C,D)}\left[\log \frac{Q(z_t|z_\lt,\tC_{\leq t},C,D)}{P(z_t|z_\lt,\tC_{\leq t},C_{\leq t})}\right] &\\
  &= \sum_{t=1}^T\eE_{\pd{t'}{t}Q(z_{t'}|z_{<t'},\tC_{\leq t'},C,D)}\left[\log P(y_t|x_t,z_t,\barC_t)\right]&\\
  &\quad- \eE_{\pd{t'}{t-1}Q(z_{t'}|z_{<t'},\tC_{\leq t'},C,D)}\mathbb{KL}\left[(Q(z_t|z_\lt,\tC_{\leq t},C,D)\parallel P(z_t|z_\lt,\tC_{\leq t},C_{\leq t}))\right] &\\
  &= \sm{t}{T} \eE_{Q_\phi(z_t|C,D)} \left[\log P_\ta(y_t|x_t,z_t,\tC_t,C_t)\right] \nn &\\
&\quad- \eE_{Q_\phi(z_\lt)}\left[\KL \left(Q_\phi(z_t|z_\lt, \tC_{\leq t}, C,D) \parallel P_\ta(z_t|z_\lt,\tC_{\leq t},C_{\leq t})) \right) \right]
\end{aligned}
\end{flalign*} where $Q_\phi(z_t|C,D)=\pd{t'}{t}Q(z_{t'}|z_{<t'},\tC_{\leq t'},C,D)$ and $Q_\phi(z_\lt)=\pd{t'}{t-1}Q(z_{t'}|z_{<t'},\tC_{\leq t'},C,D)$ for simplicity.

% to fast compile
\newpage
\FloatBarrier
\section{Qualitative Results}
\subsection{Dynamic 1D regression} \label{app:gp_res}

\begin{figure}[h!]
    \centering
        \includegraphics[width=0.75\linewidth]{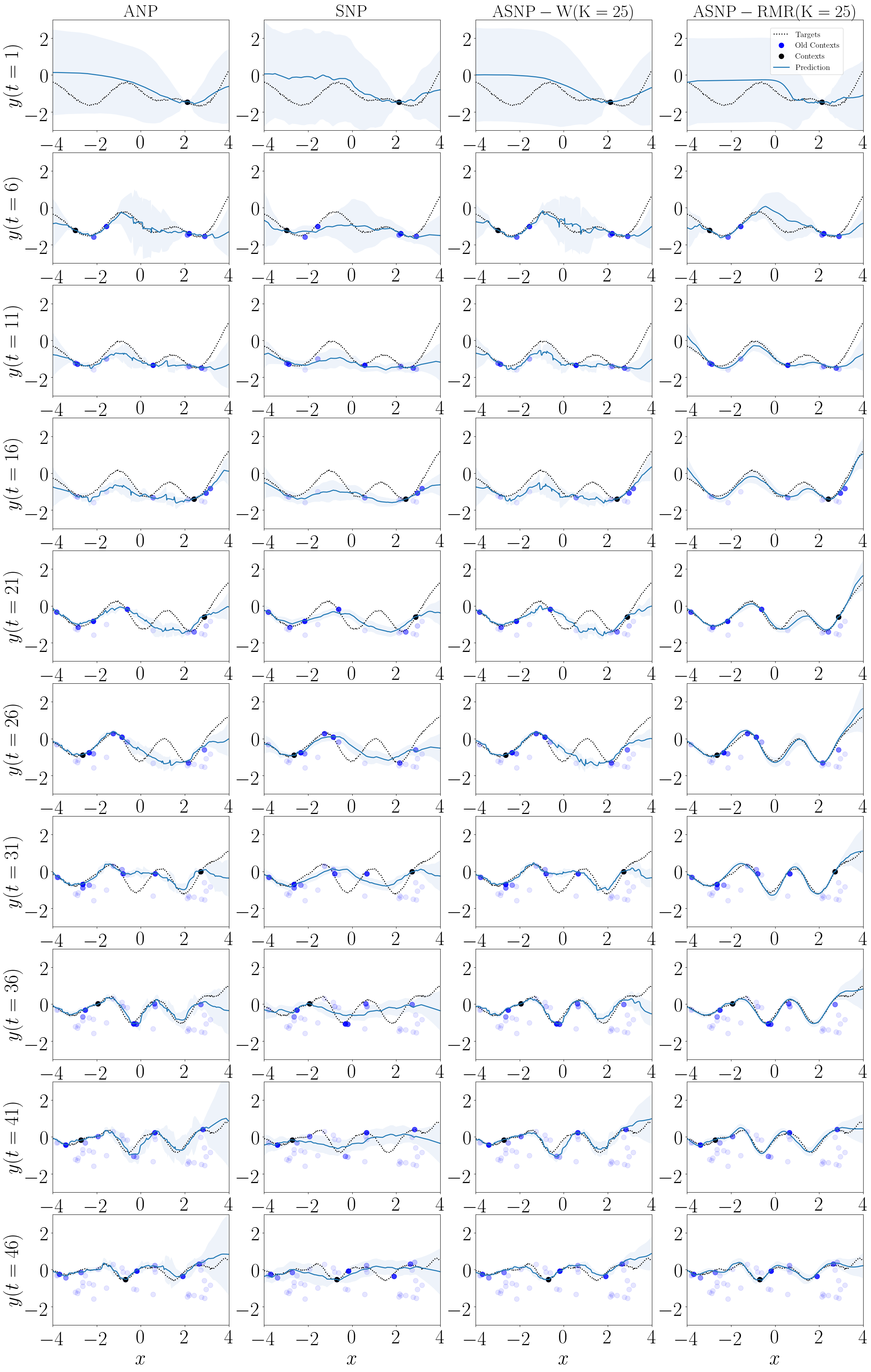}
    \caption{Dynamic 1D regression samples in sparse-context regime. Columns are ANP, SNP, ASNP-W and ASNP-RMR, respectively. Each row shows examples at each task-step. Every $5^{\text{th}}$ task-step is shown.}
    \label{app:1d_sc}
\end{figure}

\newpage

\subsection{Dynamic 2D image completion} \label{app:2d_res}
\begin{figure}[h!]
    \centering
        \includegraphics[width=0.36\linewidth]{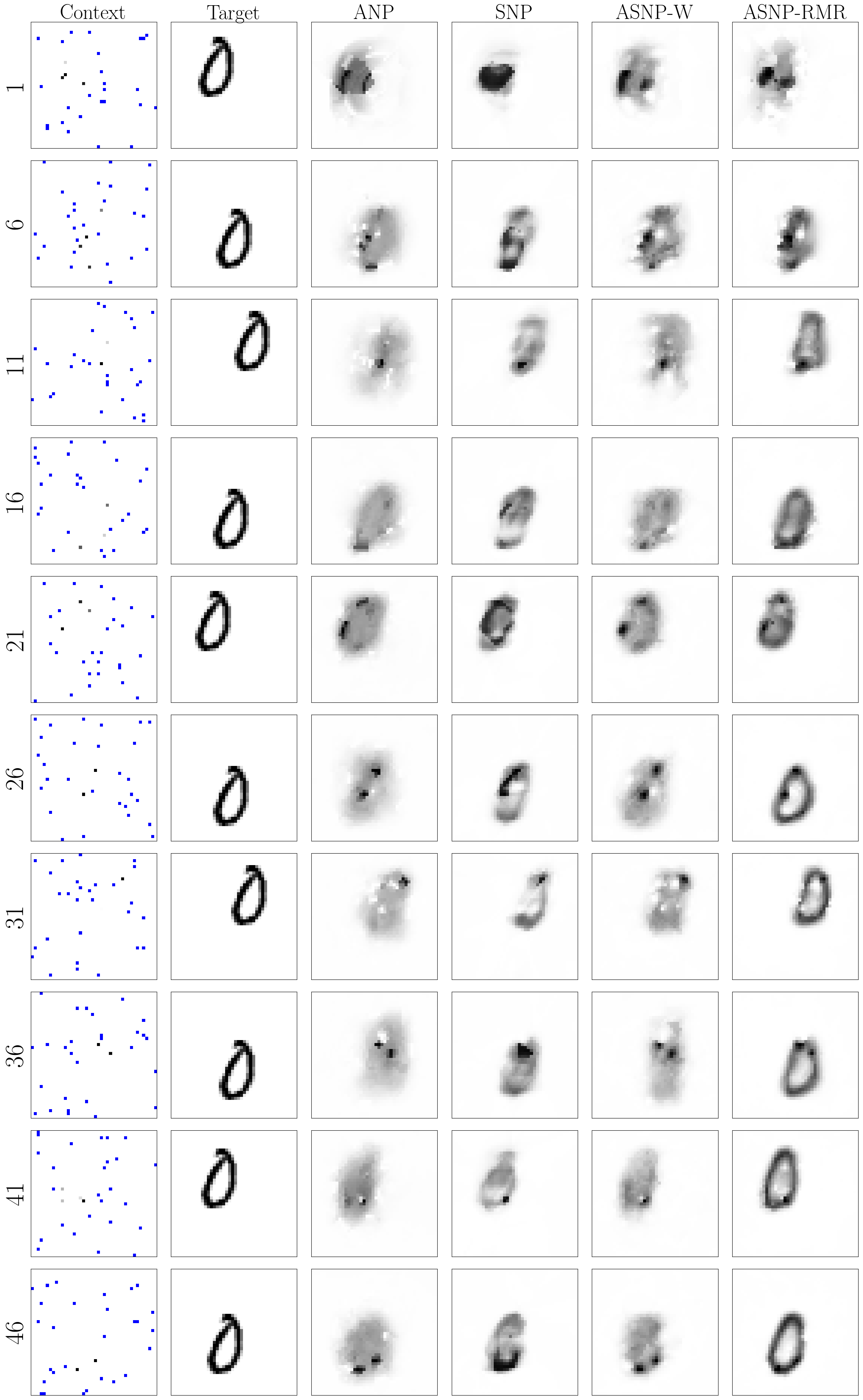}
        \includegraphics[width=0.36\linewidth]{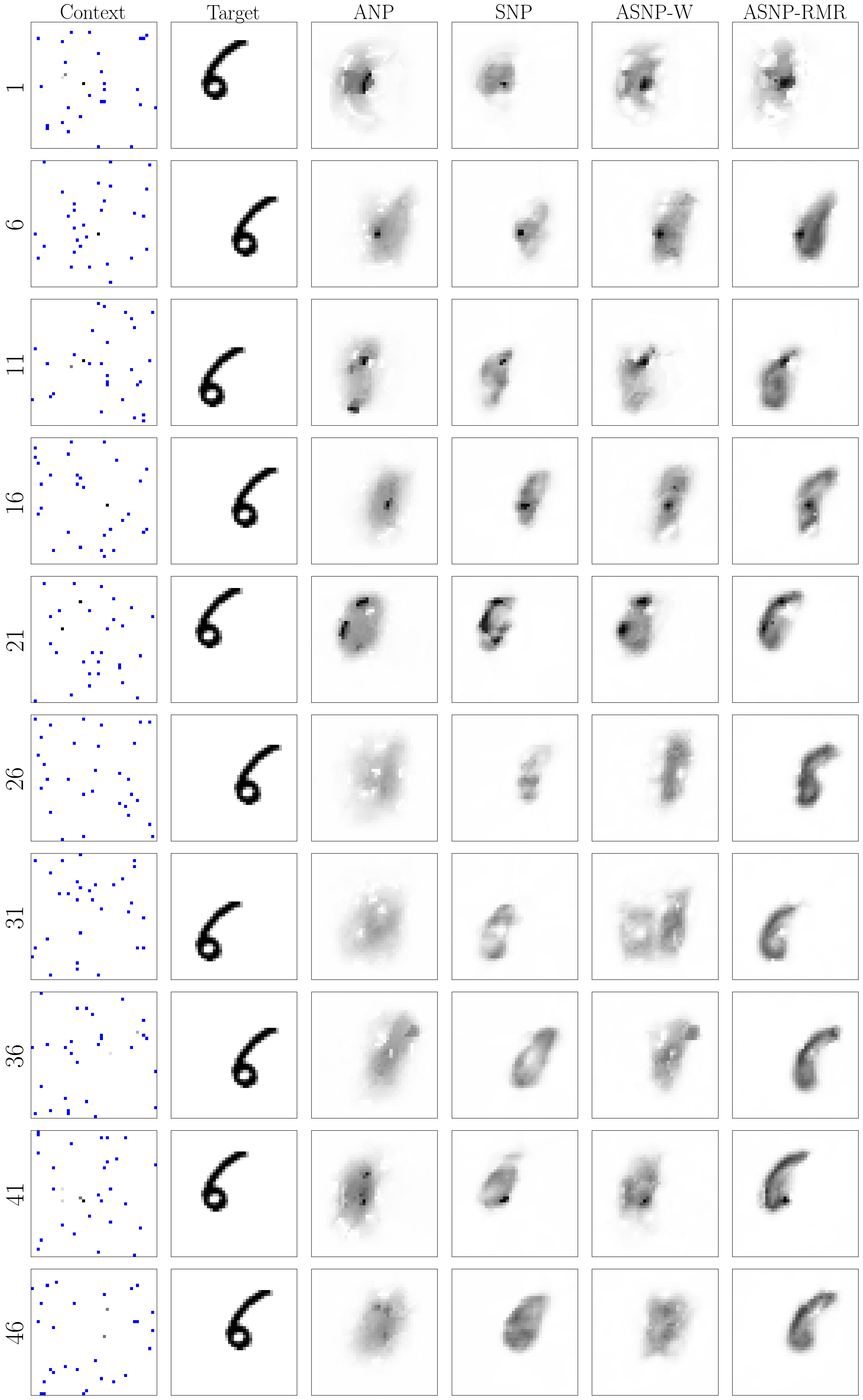}
        \includegraphics[width=0.36\linewidth]{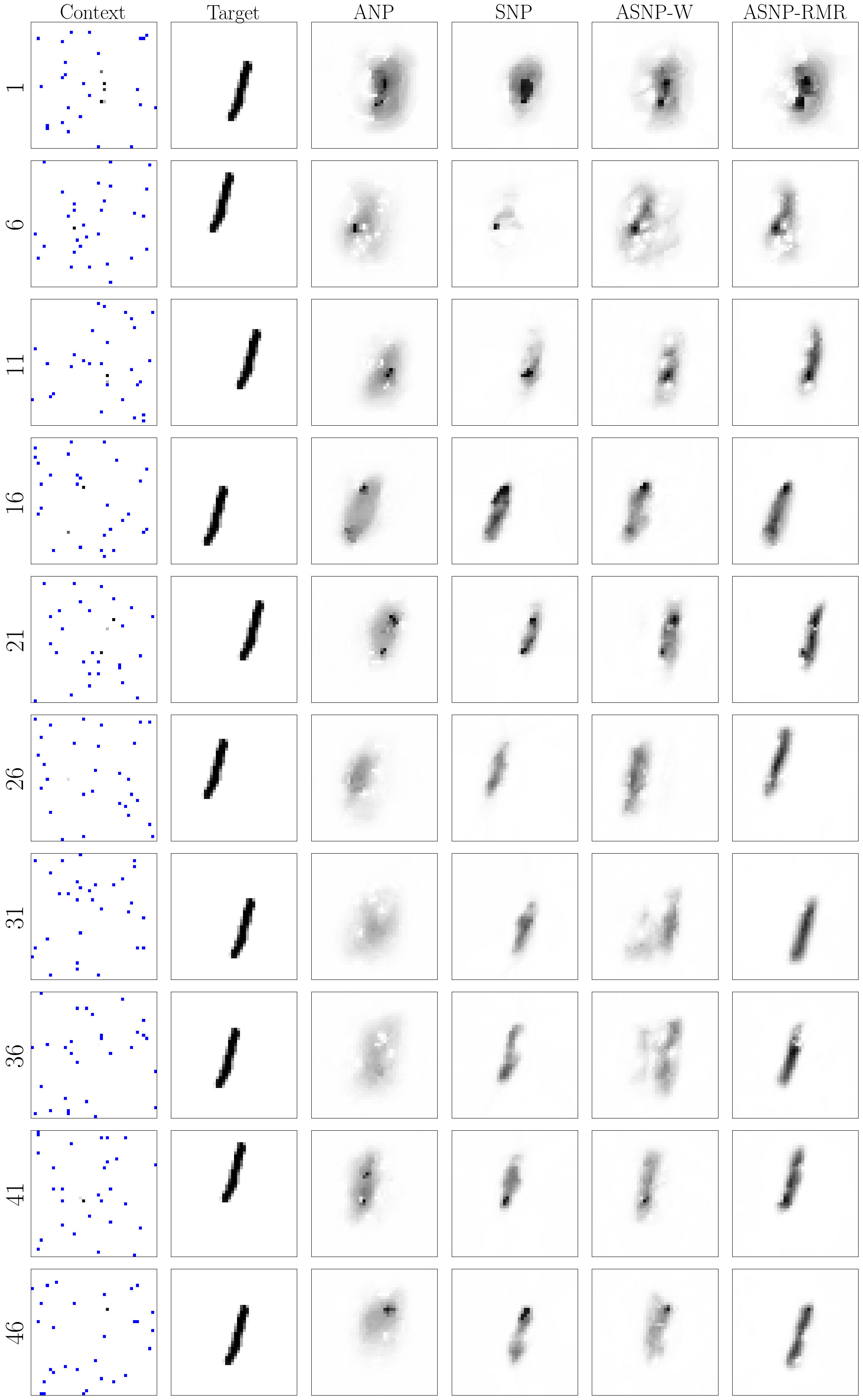}
        \includegraphics[width=0.36\linewidth]{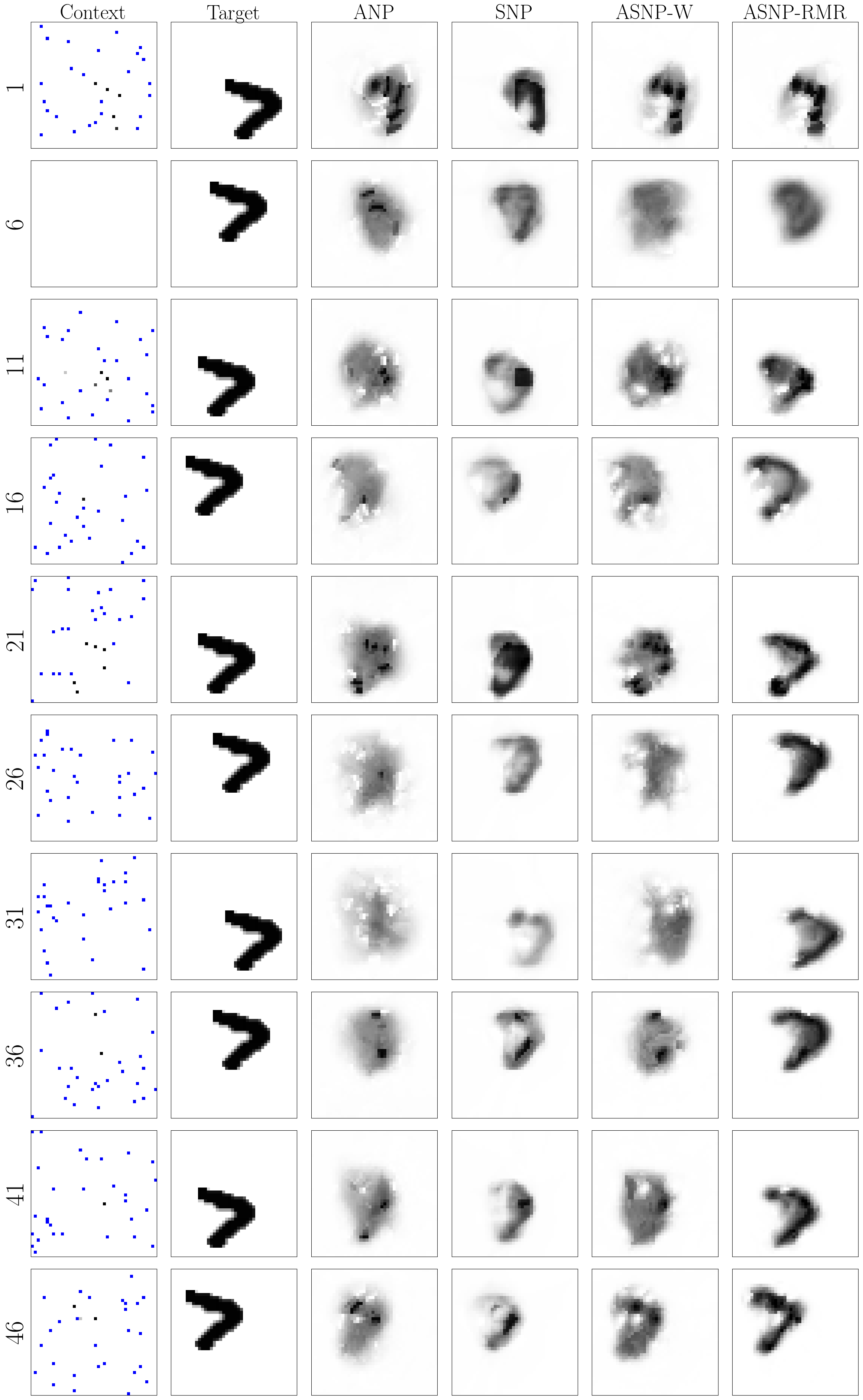}
    \caption{Dynamic moving MNIST completion samples in sparse-context regime. Columns are Context, Target, ANP, SNP, ASNP-W and ASNP-RMR, respectively. Each row shows examples at different task-steps. We show every $5^{\text{th}}$ task-step.}
    \label{app:2d_mnist_sc}
\end{figure}

\newpage

\begin{figure}[h!]
    \centering
        \includegraphics[width=0.39\linewidth]{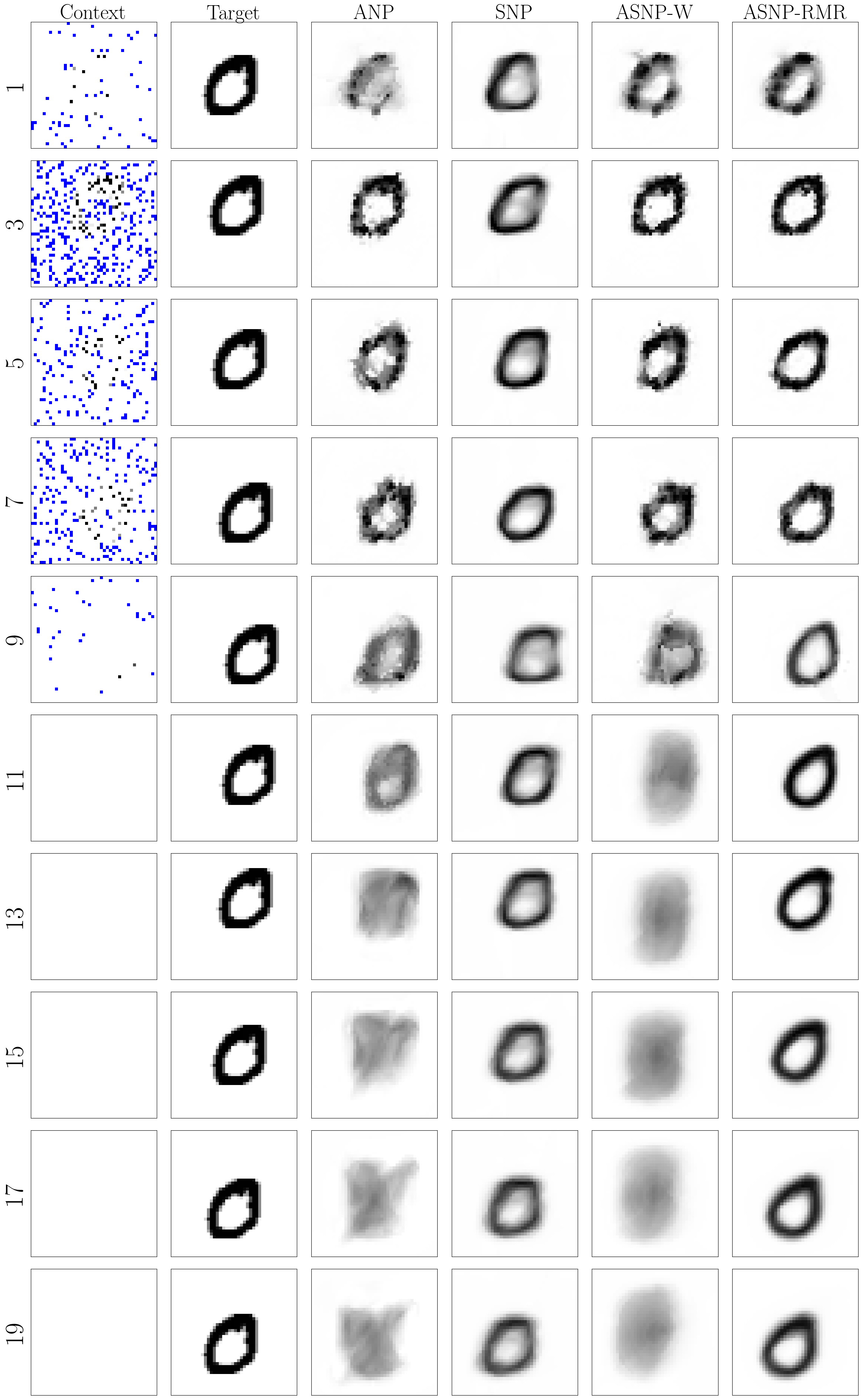}
        \includegraphics[width=0.39\linewidth]{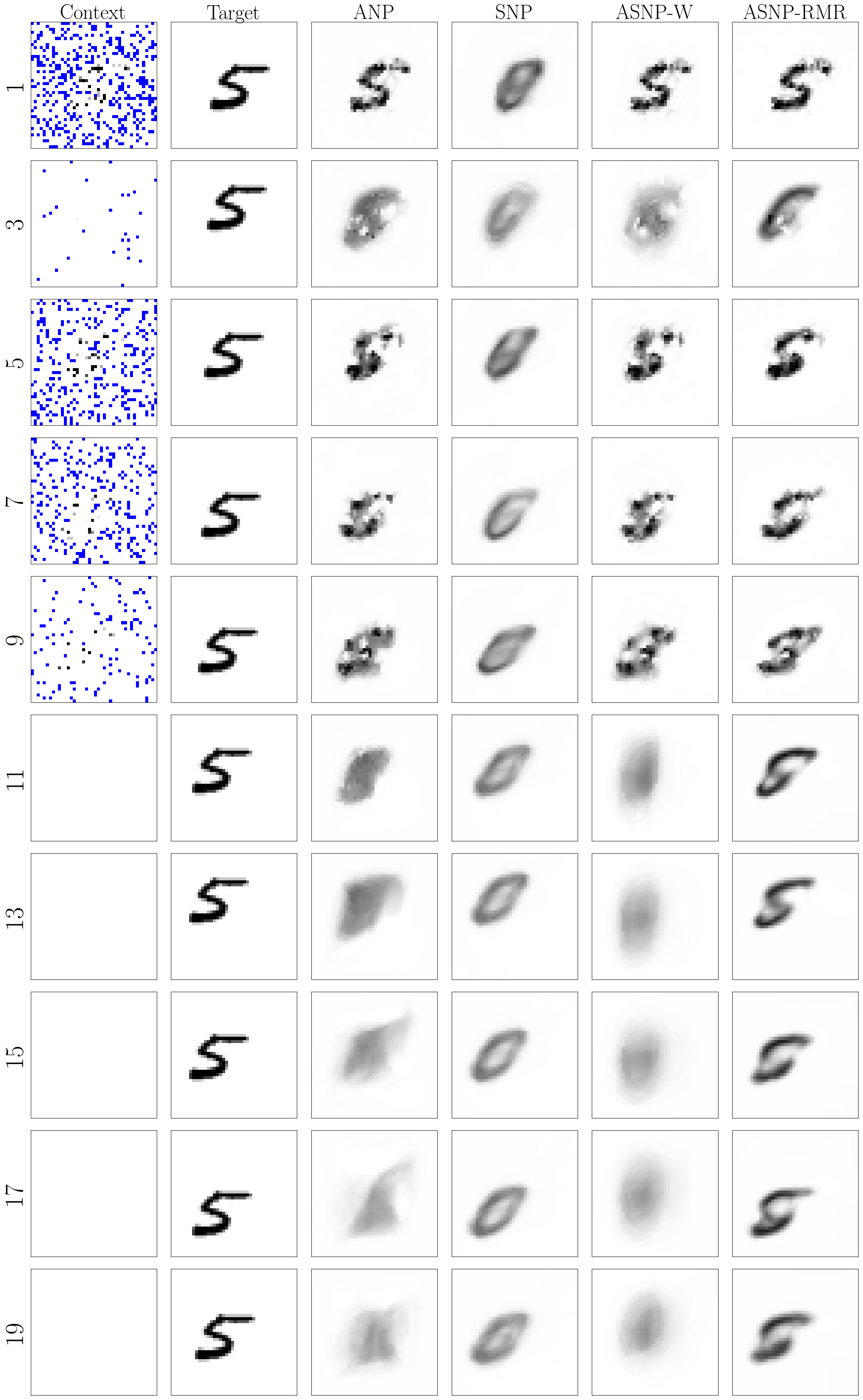}
        \includegraphics[width=0.39\linewidth]{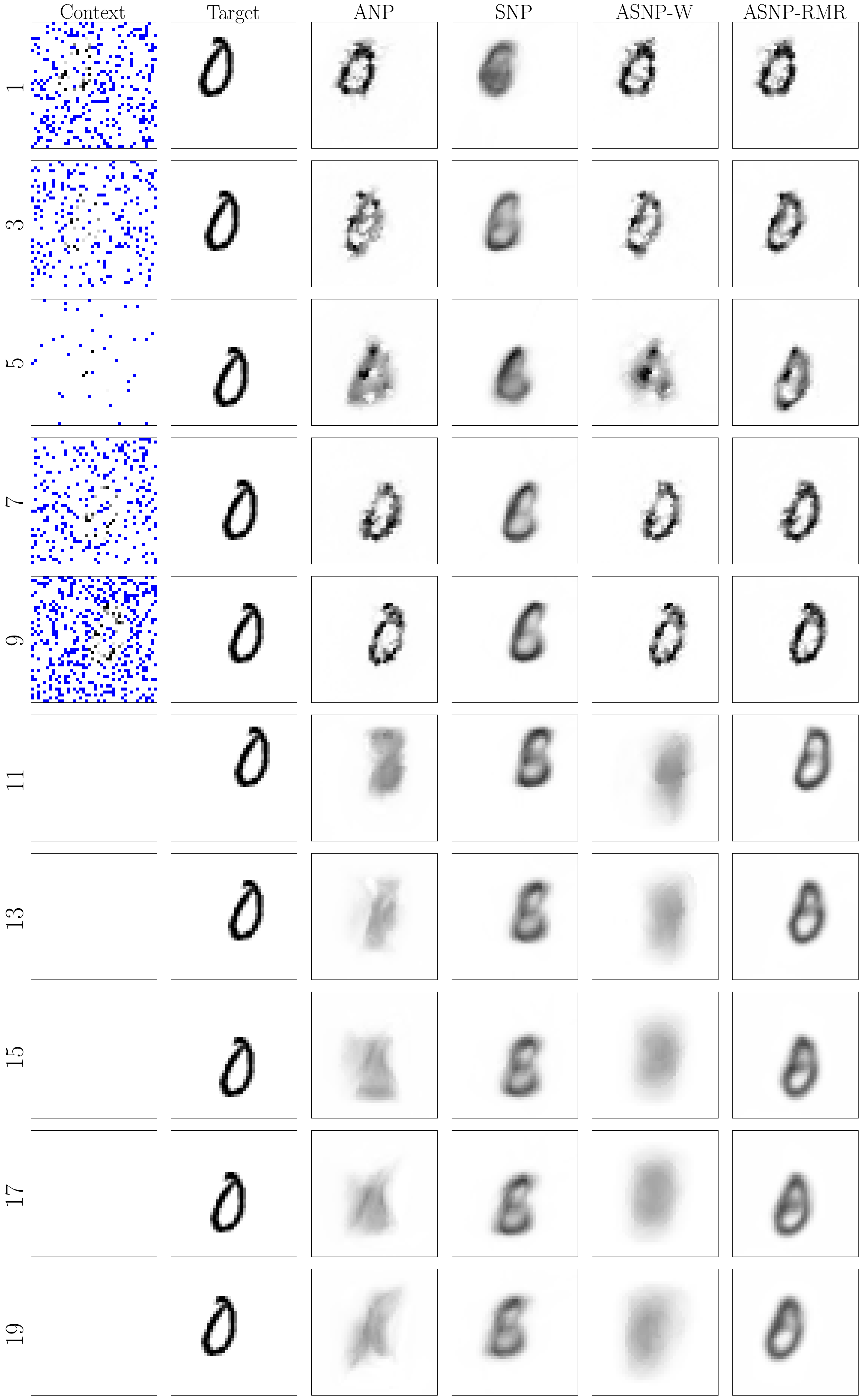}
        \includegraphics[width=0.39\linewidth]{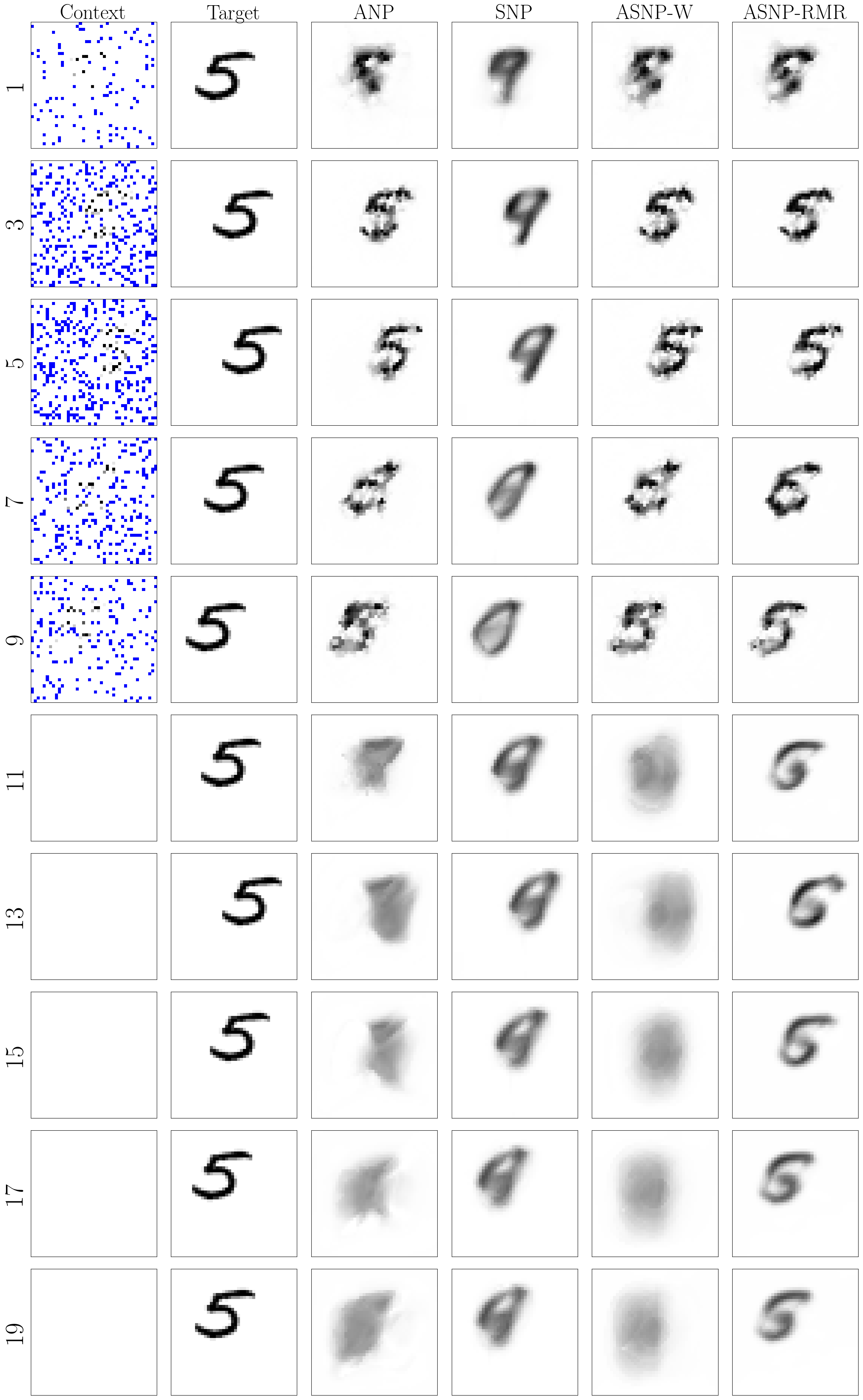}
    \caption{Dynamic moving MNIST completion samples for  transfer-prediction. Columns are Context, Target, ANP, SNP, ASNP-W and ASNP-RMR, respectively. Each row shows examples at different task-steps. We show every $2^{\text{nd}}$ task-step.}
    \label{app:2d_mnist_tp}
\end{figure}

\newpage

\begin{figure}[h!]
    \centering
        \includegraphics[width=0.39\linewidth]{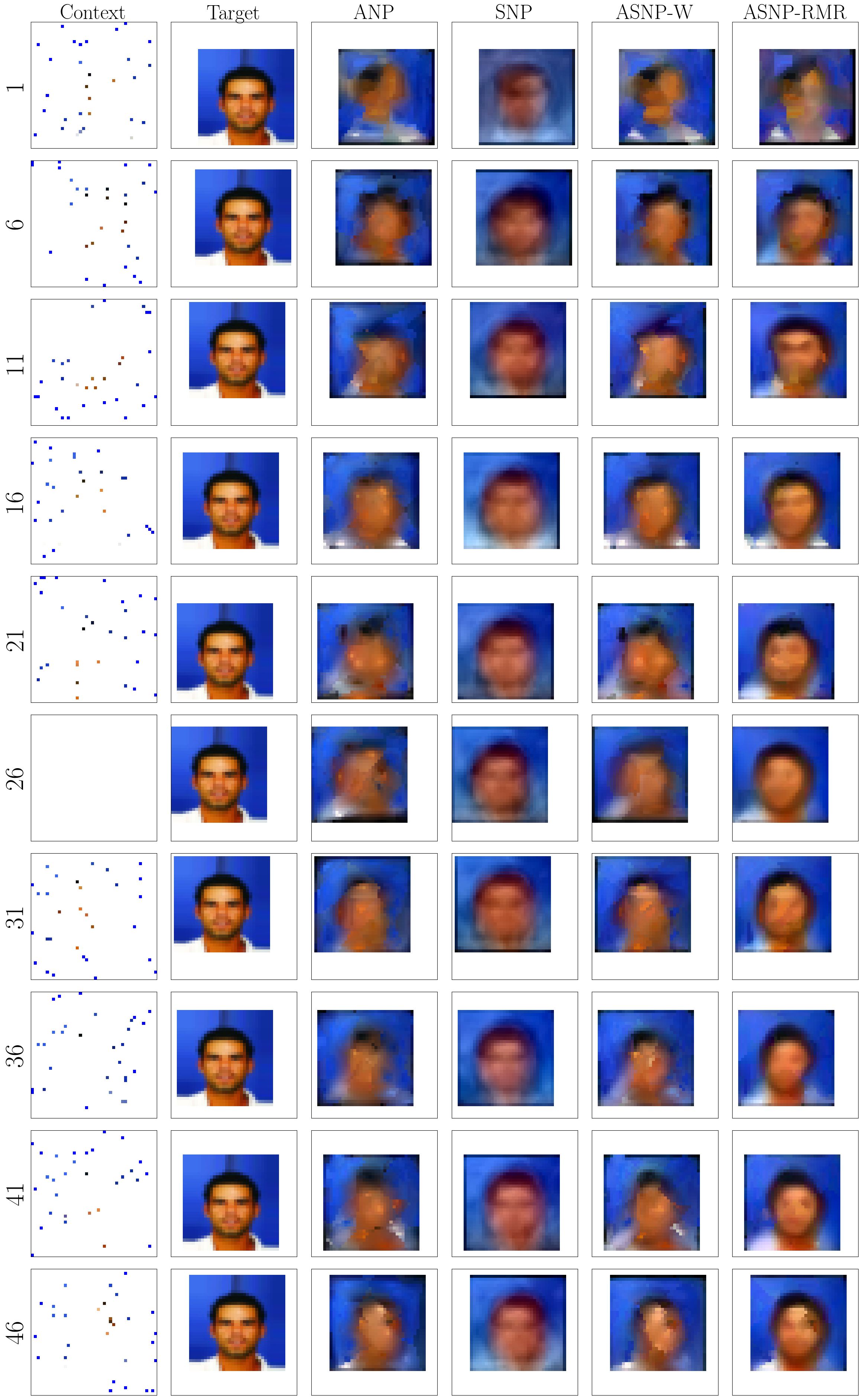}
        \includegraphics[width=0.39\linewidth]{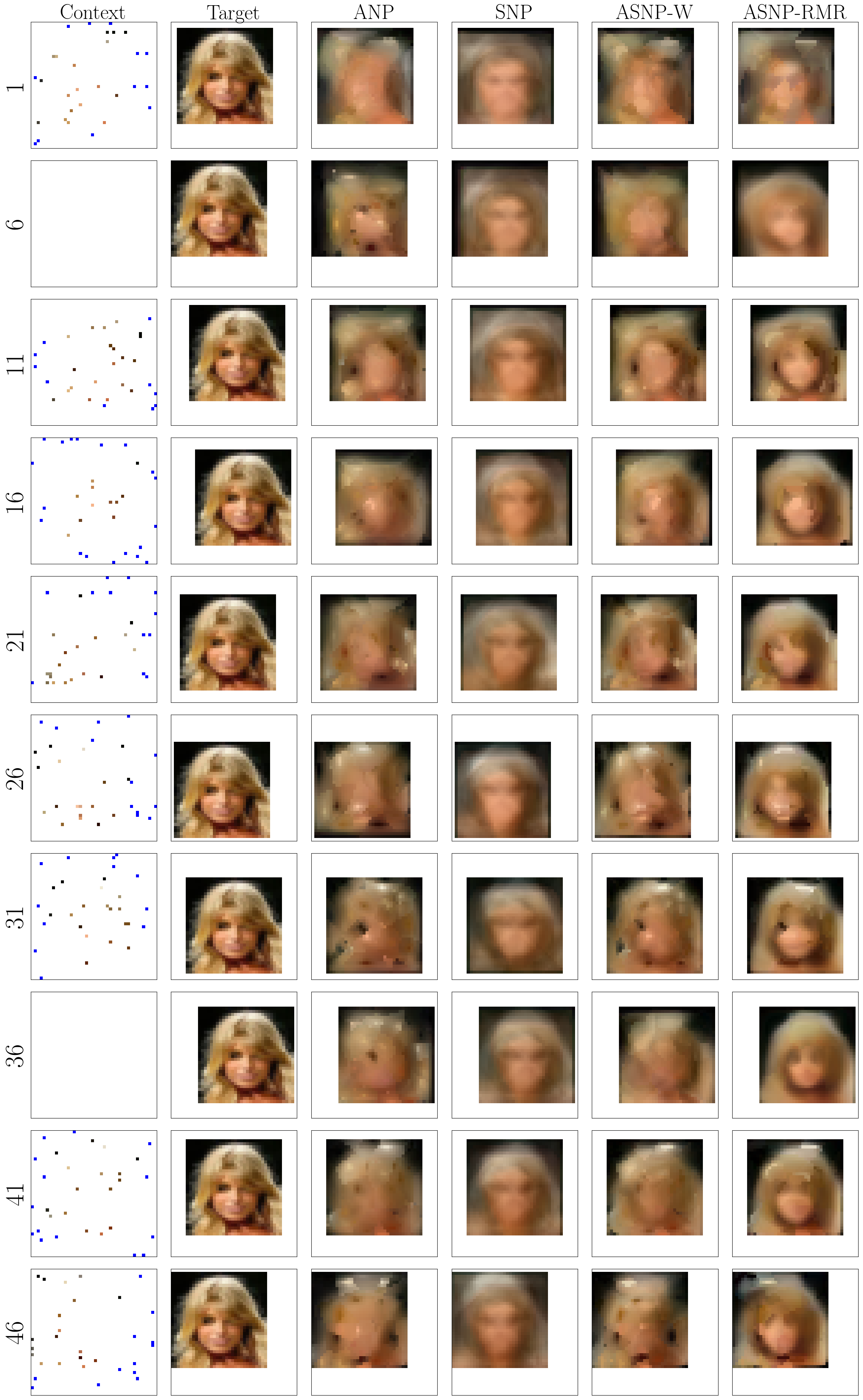}
        \includegraphics[width=0.39\linewidth]{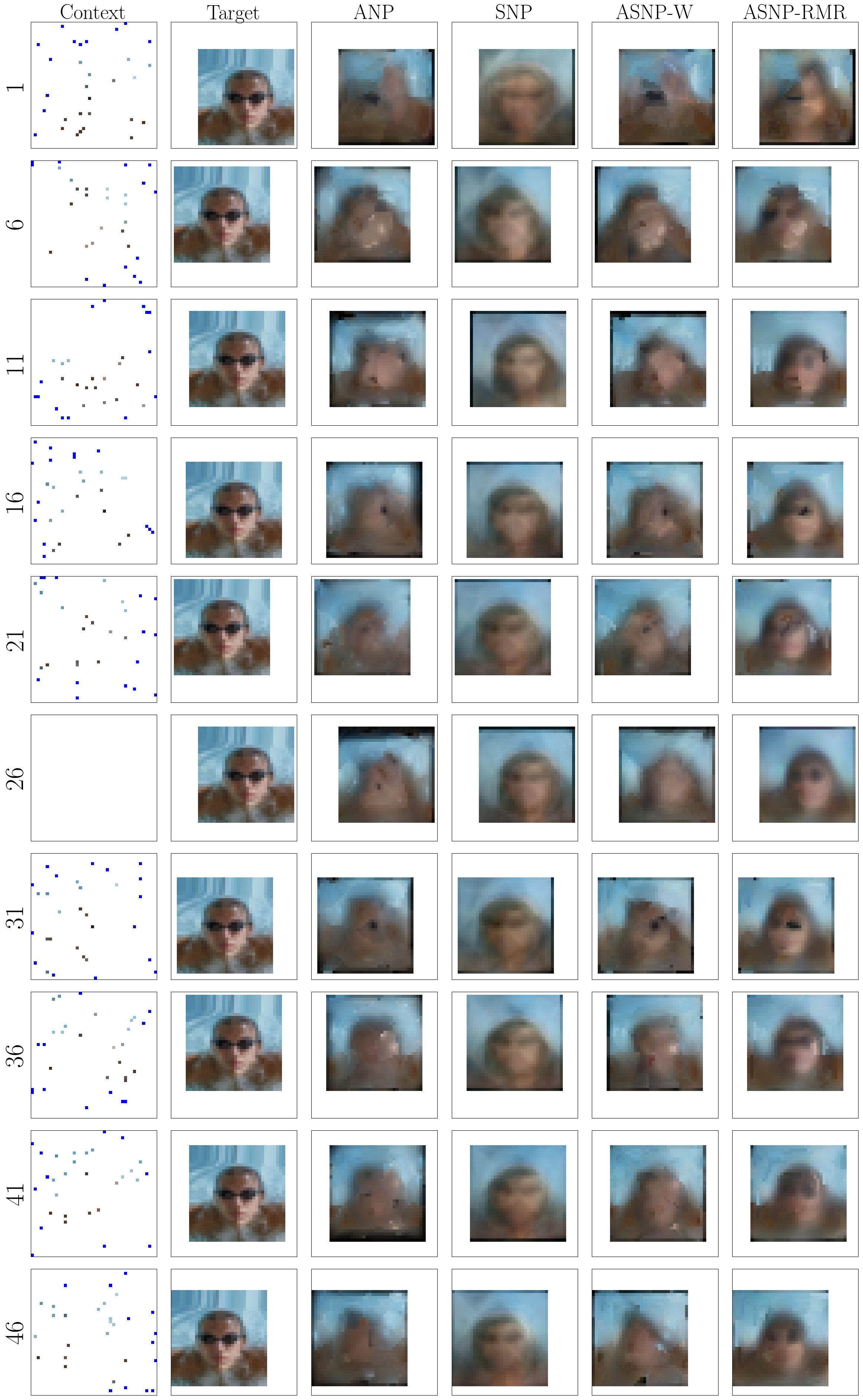}
        \includegraphics[width=0.39\linewidth]{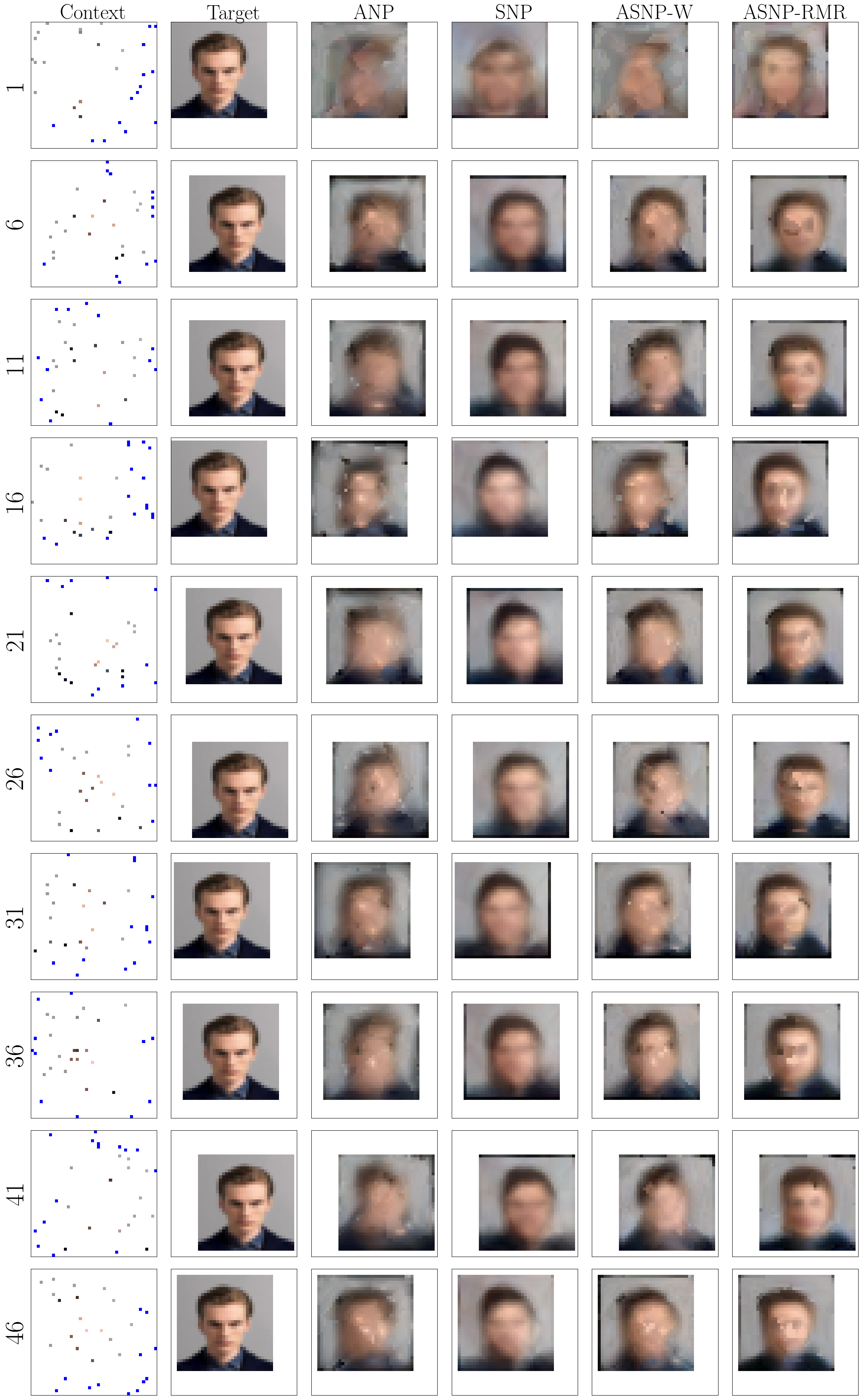}
    \caption{Dynamic moving CelebA completion samples in sparse-context regime. Columns are Context, Target, ANP, SNP, ASNP-W and ASNP-RMR, respectively. Each row shows examples at each task-step. Every $5^{\text{th}}$ task-step is shown here.}
    \label{app:2d_celeba_c3}
\end{figure}

\newpage

\begin{figure}[h!]
    \centering
        \includegraphics[width=0.39\linewidth]{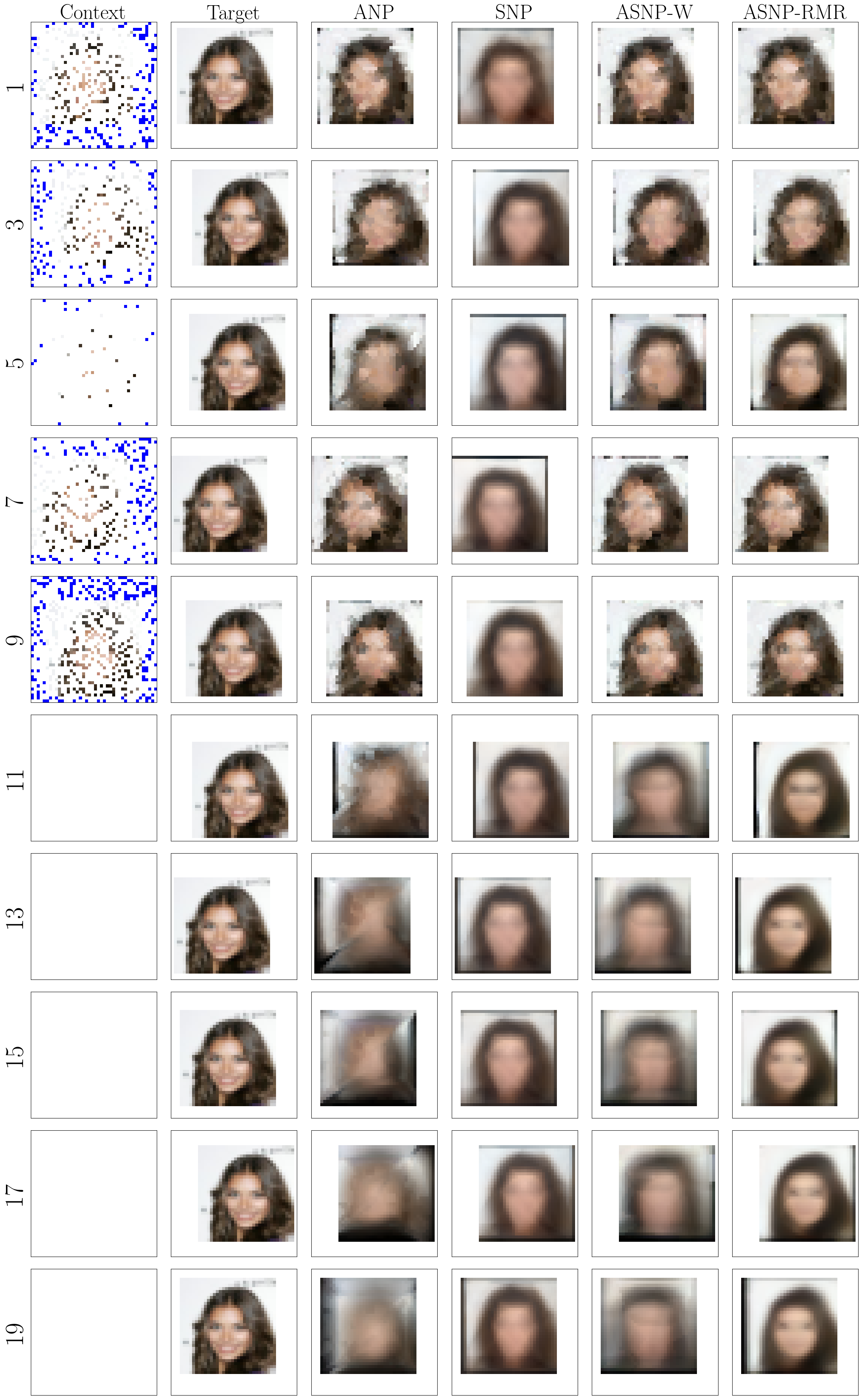}
        \includegraphics[width=0.39\linewidth]{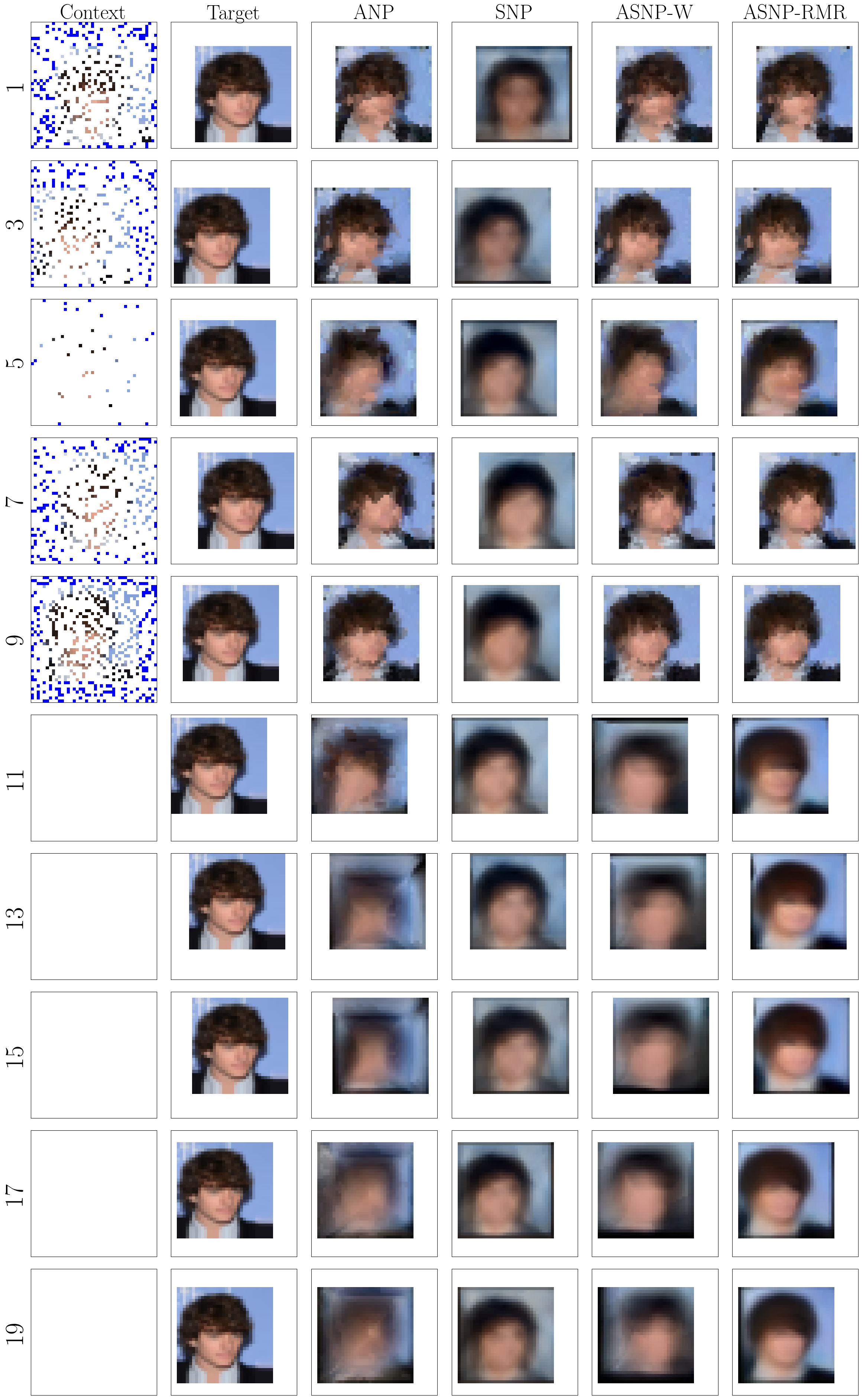}
        \includegraphics[width=0.39\linewidth]{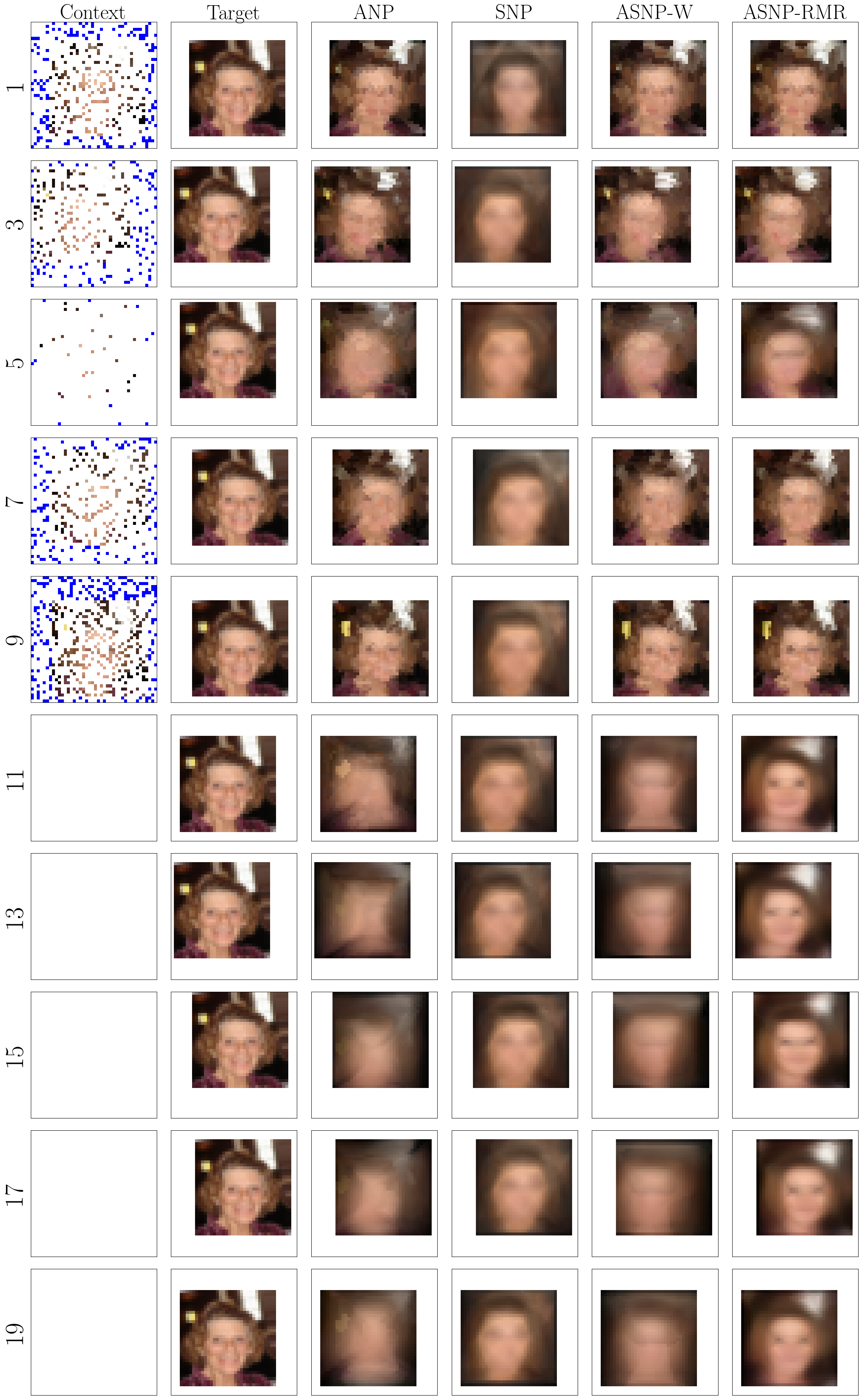}
        \includegraphics[width=0.39\linewidth]{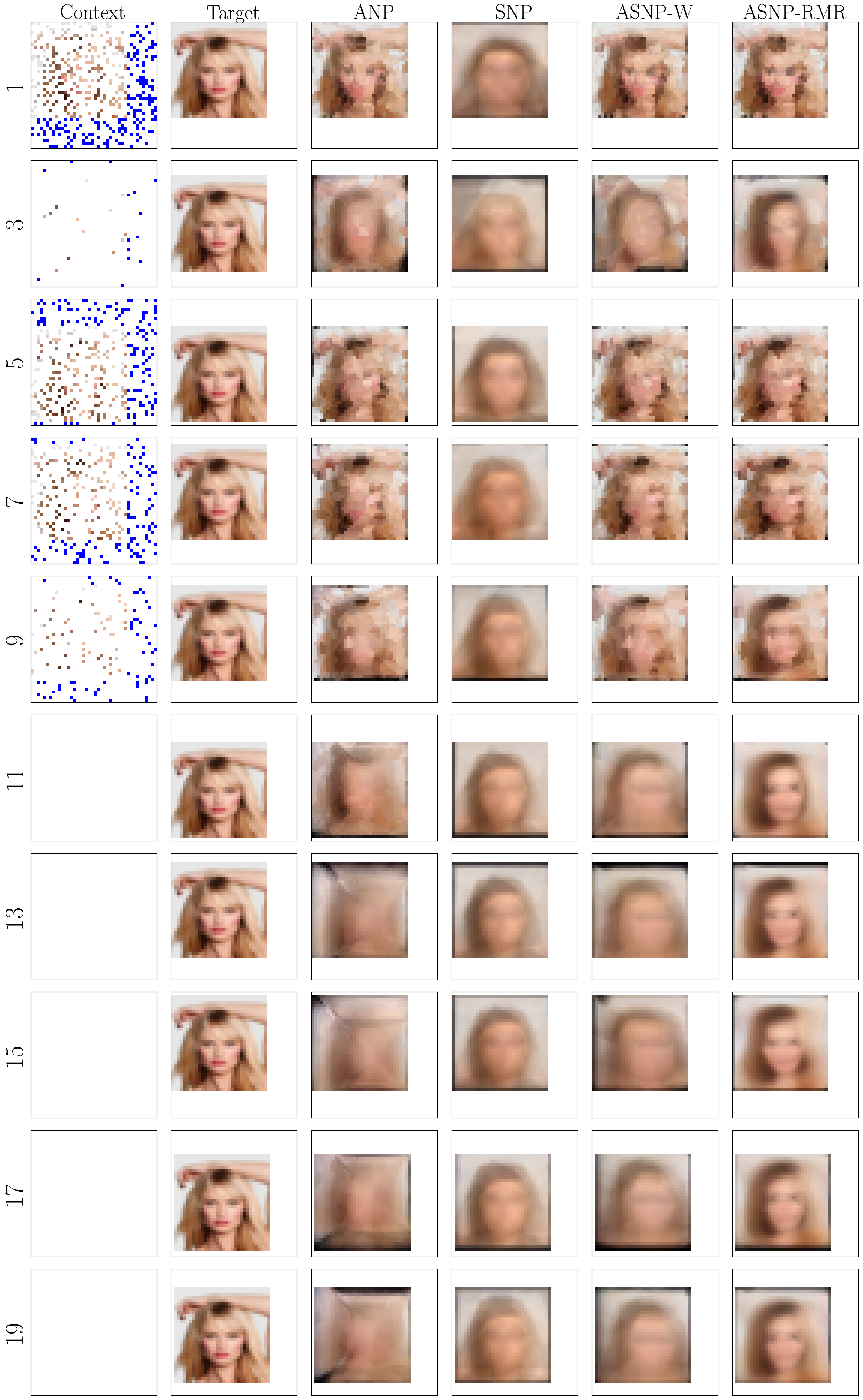}
    \caption{Dynamic moving CelebA completion samples for transfer-prediction. Columns are Context, Target, ANP, SNP, ASNP-W and ASNP-RMR, respectively. Each row shows examples at each task-step. Every $2^{\text{nd}}$ task-step is shown here.}
    \label{app:2d_celeba_c1}
\end{figure}

\end{appendices}

\end{document}